\documentclass[sn-mathphys]{sn-jnl}
\jyear{2021}%

\theoremstyle{thmstyleone}%
\usepackage{framed}
\usepackage{listings}
\usepackage{soul}

\usepackage{booktabs} 
\usepackage{arydshln}

\theoremstyle{thmstyletwo}%

\theoremstyle{thmstylethree}%

\usepackage{amsmath,amssymb} 
\usepackage{cleveref}
\usepackage{color}
\usepackage{multirow}
\usepackage{graphicx}
\usepackage{comment}
\usepackage{changepage}
\usepackage{wrapfig}
\usepackage{colortbl} 
\usepackage{arydshln} 
\usepackage{amsthm}
\usepackage{bm}
\usepackage{rotating}
\usepackage{amsmath}

\definecolor{darkgreen}{RGB}{5,102,8}

\begin{document}

\title[ ]{From reactive to cognitive: brain-inspired spatial intelligence for embodied agents}

%%=============================================================%%

\author[1,2]{\fnm{Shouwei} \sur{Ruan}}\email{shouweiruan@buaa.edu.cn}
\equalcont{These authors contributed equally to this work.}

\author[1,3]{\fnm{Liyuan} \sur{Wang}}\email{liyuanwang@tsinghua.edu.cn}
\equalcont{These authors contributed equally to this work.}

\author[2]{\fnm{Caixin} \sur{Kang}}\email{caixinkang@buaa.edu.cn}

\author[2]{\fnm{Qihui} \sur{Zhu}}\email{sy2202227@buaa.edu.cn}

\author[1]{\fnm{Songming} \sur{Liu}}\email{lsm23@mails.tsinghua.edu.cn}

\author*[2]{\fnm{Xingxing} \sur{Wei}}\email{xxwei@buaa.edu.cn}

\author*[1]{\fnm{Hang} \sur{Su}}\email{suhangss@mail.tsinghua.edu.cn}

\affil[1]{\orgdiv{Department of Computer Science and Technology, Institute for AI, BNRist Center, Tsinghua-Bosch Joint ML Center, THBI Lab}, \orgname{Tsinghua University}, \orgaddress{\state{Beijing}, \country{China}}}

\affil[2]{\orgdiv{Institute of Artificial Intelligence}, \orgname{Beihang University}, \orgaddress{\state{Beijing}, \country{China}}}

\affil[3]{\orgdiv{Department of Psychological and Cognitive Sciences}, \orgname{Tsinghua University}, \orgaddress{\state{Beijing}, \country{China}}}


%150 words
\abstract{
Spatial cognition enables adaptive goal-directed behavior by constructing internal models of space. Robust biological systems consolidate spatial knowledge into three interconnected forms: \textit{landmarks} for salient cues, \textit{route knowledge} for movement trajectories, and \textit{survey knowledge} for map-like representations. While recent advances in multi-modal large language models (MLLMs) have enabled visual-language reasoning in embodied agents, these efforts lack structured spatial memory and instead operate reactively, limiting their generalization and adaptability in complex real-world environments. Here we present Brain-inspired Spatial Cognition for Navigation (BSC-Nav), a unified framework for constructing and leveraging structured spatial memory in embodied agents. BSC-Nav builds allocentric cognitive maps from egocentric trajectories and contextual cues, and dynamically retrieves spatial knowledge aligned with semantic goals. Integrated with powerful MLLMs, BSC-Nav achieves state-of-the-art efficacy and efficiency across diverse navigation tasks, demonstrates strong zero-shot generalization, and supports versatile embodied behaviors in the real physical world, offering a scalable and biologically grounded path toward general-purpose spatial intelligence. Our code is available at \href{https://github.com/Heathcliff-saku/BSC-Nav}{}, and the Supplementary video is available at \href{Google_Drive}{https://drive.google.com/drive/folders/1p1GjpQMQQ-ylmazhjPgqT49AUOcUT3Z-?usp=sharing}
}




\maketitle

%3500 words (intro+result+discussion)
\section{Introduction}\label{sec1}
%%%%%%%%%%%%%%%%%%%%%%%%%%%%%%%%%

Spatial cognition, the ability to acquire, organize, exploit, and update knowledge about external space, is fundamental to both human beings and artificial intelligence (AI)~\cite{gupta2021embodied,malanchini2020evidence}. 
It underlies not only sensorimotor skills such as navigation and manipulation, but also supports higher-level cognitive functions, including abstraction, planning, and reasoning. In humans, generalizable spatial representations support interpretation of sensory input, anticipation of future events, and flexible adaptation to changing environments, from brewing coffee in a familiar kitchen to navigating an unfamiliar city~\cite{bellmund2018navigating, epstein2017cognitive, denis2007perspectives}. Given its broad relevance in embodied interaction with the real physical world, spatial cognition has emerged as a central theme across disciplines, driving advances in robotics~\cite{yang2025thinking}, urban simulation~\cite{tucker2024systematic}, and planetary-scale modeling~\cite{papadimitriouspatial}, while increasingly recognized as a foundational component of artificial general intelligence (AGI)~\cite{bellmund2018navigating,banino2018vector,bermudez2020neuroscience}.

Despite the rapid progress in multi-modal large language models (MLLMs) and growing efforts to equip embodied agents with visual-language reasoning~\cite{cai2024bridging, zhang2024uni, yokoyama2024vlfm}, current artificial systems remain fundamentally limited in large-scale spatial cognition~\cite{ramakrishnan2024does}, especially in tasks requiring long-horizon navigation and mobile manipulation. A central bottleneck lies in the lack of \textbf{structured spatial memory}~\cite{mcnamara1989subjective,werner1997spatial,siegel1975development}, a mechanism for persistently encoding, organizing, and retrieving spatial knowledge about the environments.
Most existing methods, whether based on end-to-end reinforcement learning~\cite{chaplot2020object, yadav2023offline, shah2023vint} or modular pipelines with powerful MLLMs~\cite{openai2023gpt4v_systemcard, yang2023qwenvl}, process observations in a reactive and stateless manner. Without durable internal models of external space, agents struggle to consolidate coherent spatial representations or perform reasoning beyond immediate stimuli, resulting in fragmented knowledge, short-sighted planning, and poor generalization. Addressing these limitations requires a paradigm shift from reactive processing to memory-centric spatial cognition, which supports persistent representations and compositional reasoning over time and space.

\begin{figure*}[t]
  \centering
  \includegraphics[width=0.98\linewidth]{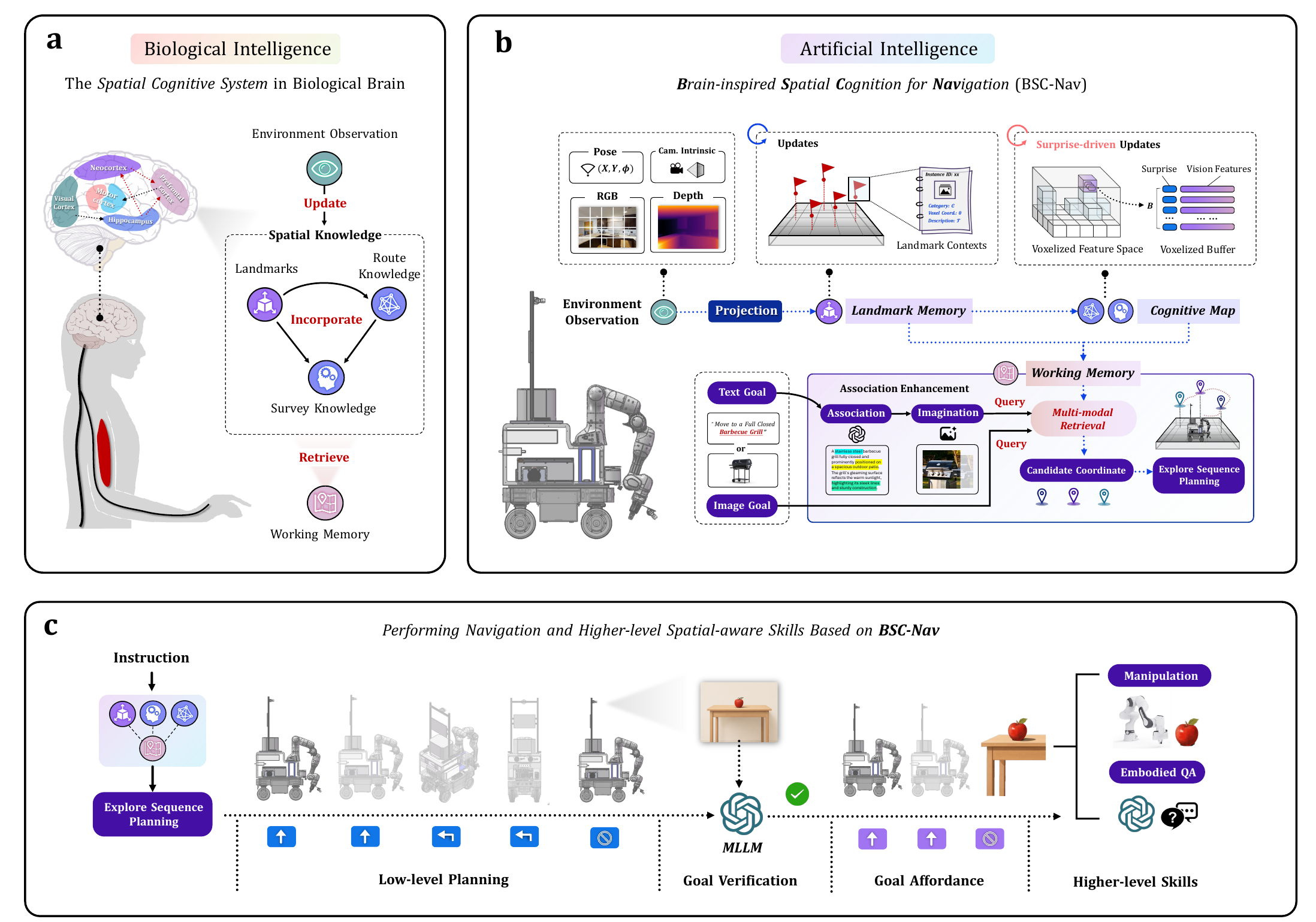}
   \caption{
   %\textbf{Biological spatial cognition principles inspire the design of the BSC-Nav framework.} 
   \textbf{The BSC-Nav framework for cognitive spatial intelligence.}
   \textbf{a}, Structured spatial memory in biological brains, comprising landmarks, route knowledge, and survey knowledge.
   \textbf{b}, The BSC-Nav framework instantiates structured spatial memory in embodied agents. Environment observations (RGB-D images and agent poses) are processed by (i) a landmark memory module encodes and retrieves durable associations of multi-modal environmental cues as the \textit{landmarks}; and (ii) a cognitive map module accumulates and organizes movement trajectories as the \textit{route knowledge} into allocentric, map-like representations as the \textit{survey knowledge}. Upon task invocation, (iii) a working memory module dynamically composes relevant spatial knowledge for adaptive planning and reasoning. %MLLMs interact flexibly with spatial knowledge to achieve robust retrieval, multi-level reasoning and precise goal localization.
   \textbf{c}, Structured spatial memory enables not only universal navigation but also higher-level spatial-aware skills. % including mobile manipulation and embodied question answering.
   }
   \label{fig:Figure1}
\end{figure*}

Compared to artificial systems, biological spatial cognition offers a robust and compelling template. Decades of neuroscience research revealed that organisms consolidate spatial knowledge into three distinct yet interconnected forms~\cite{werner1997spatial,siegel1975development} (Fig.~\ref{fig:Figure1}a): 
\textit{landmarks}, which encode stable associations of salient environmental cues to support localization and contextual understanding~\cite{richter2014landmarks,jansen2006wayfinding}; \textit{route knowledge}, which captures egocentric movement trajectories between landmarks for habitual navigation and path integration~\cite{wolbers2004neural}; and \textit{survey knowledge}, which integrates multiple routes into allocentric, map-like representations that support flexible inference, shortcut discovery, and detour planning~\cite{chrastil2013active}. These spatial representations are accessed and coordinated via working memory~\cite{baddeley2012working}, especially visual-spatial working memory~\cite{awh2001overlapping,logie2014visuo}, enabling adaptive retrieval, composition, and generalization based on task demands and environmental familiarity.

Motivated by these biological principles, we introduce Brain-inspired Spatial Cognition for Navigation (BSC-Nav), a unified framework that instantiates cognitive spatial intelligence in embodied agents via structured spatial memory (Fig.~\ref{fig:Figure1}b). 
BSC-Nav explicitly constructs spatial knowledge with two synergistic branches: a \textbf{landmark memory module}, which encodes durable associations between salient environmental cues and spatial locations to instantiate \textit{landmarks}; and a \textbf{cognitive map module}, which accumulates \textit{route knowledge} by transforming egocentric movement sequences into voxelized trajectories, and organizes them into allocentric, map-like representations as \textit{survey knowledge}. 
BSC-Nav further incorporates a \textbf{working memory module} that dynamically retrieves and combines spatial representations from landmark memory and cognitive map, thereby aligning semantic goals with grounded spatial actions.
Each component interfaces seamlessly with large-scale foundation models: visual foundation models like DINOv2~\cite{oquab2023dinov2, darcet2023vitneedreg} provide perceptual grounding of environmental cues, while MLLMs like GPT-4V~\cite{openai2023gpt4v_systemcard} guide high-level semantic interpretation and goal-conditioned reasoning. 

By integrating MLLMs-based embodied agents with structured spatial memory, BSC-Nav achieves robust spatial cognition capabilities, supporting long-horizon reasoning, experience reuse, and flexible transitions between local and global policies. BSC-Nav achieves state-of-the-art performance across a broad spectrum of navigation tasks, including object-goal, open-vocabulary, and instance-level navigation, while also demonstrating strong spatial generalization in instruction following, embodied question answering, and mobile manipulation. These results position BSC-Nav as a scalable and biologically grounded solution that complements MLLMs's vision-language reasoning with cognitive spatial intelligence, enabling more capable, adaptable, and cognitively informed AI operating in the real physical world.

\section{Results}\label{sec2}

We begin by presenting the BSC-Nav framework, which instantiates brain-inspired structured spatial memory in embodied agents. We then systematically evaluate its performance across both simulated and real-world scenarios, highlighting: (i) universal navigation capabilities in foundational tasks; (ii) higher-level spatially-ware skills in instruction-driven visual-language navigation and active embodied question answering; and (iii) real-world efficacy in navigation and mobile manipulation.

\subsection{Construction and exploitation of structured spatial memory in embodied agents}

Structured spatial memory is essential for guiding goal-directed behavior in complex real-world environments~\cite{bellmund2018navigating, epstein2017cognitive, denis2007perspectives}, yet remains largely absent from current embodied agents. Drawing inspiration from the robust spatial representations in biological systems, namely, \textit{landmarks}~\cite{richter2014landmarks, jansen2006wayfinding}, \textit{route knowledge}~\cite{wolbers2004neural}, and \textit{survey knowledge}~\cite{chrastil2013active}, BSC-Nav implements a modular architecture that explicitly constructs and exploits analogous memory structures to achieve spatial cognition capabilities (Fig.~\ref{fig:Figure1}b, detailed in Methods).

BSC-Nav comprises two branches that are continuously updated to accumulate and organize spatial knowledge during \textbf{environment exploration}.
First, a \textbf{landmark memory module} encodes salient environmental cues as associative triplets comprising spatial coordinates, semantic categories, and contextual descriptions, consolidating \textit{landmarks} of the surroundings. This design yields abstract and sparse representations that prioritize salient instances, enabling efficient retrieval and forming a flexible, interpretable scaffold of the external space (detailed in Methods).
Meanwhile, a \textbf{cognitive map module} transforms egocentric observations along movement trajectories into \textit{route knowledge}, which are then voxelized into persistent allocentric representations as \textit{survey knowledge} (detailed in Methods). Inspired by the free-energy principle~\cite{friston2010free}, which posits that biological systems minimize prediction error to refine internal models~\cite{friston2010free,friston2009free}, we implement a surprise-driven update strategy that selectively integrates novel or unexpected spatial observations. Additionally, a voxelized memory buffer maintains diverse spatial representations across viewpoints and timepoints, enhancing robustness and generalization.

To exploit structured spatial memory during \textbf{task execution}, BSC-Nav further incorporates a \textbf{working memory module} that adaptively retrieves landmarks, route knowledge, and survey knowledge for goal localization and trajectory planning (detailed in Methods). Analogous to visual-spatial working memory in biological systems~\cite{awh2001overlapping, logie2014visuo}, this module dynamically composes spatial representations based on task demands and environmental familiarity. 

\begin{figure*}[t]
  \centering
  \includegraphics[width=0.98\linewidth]{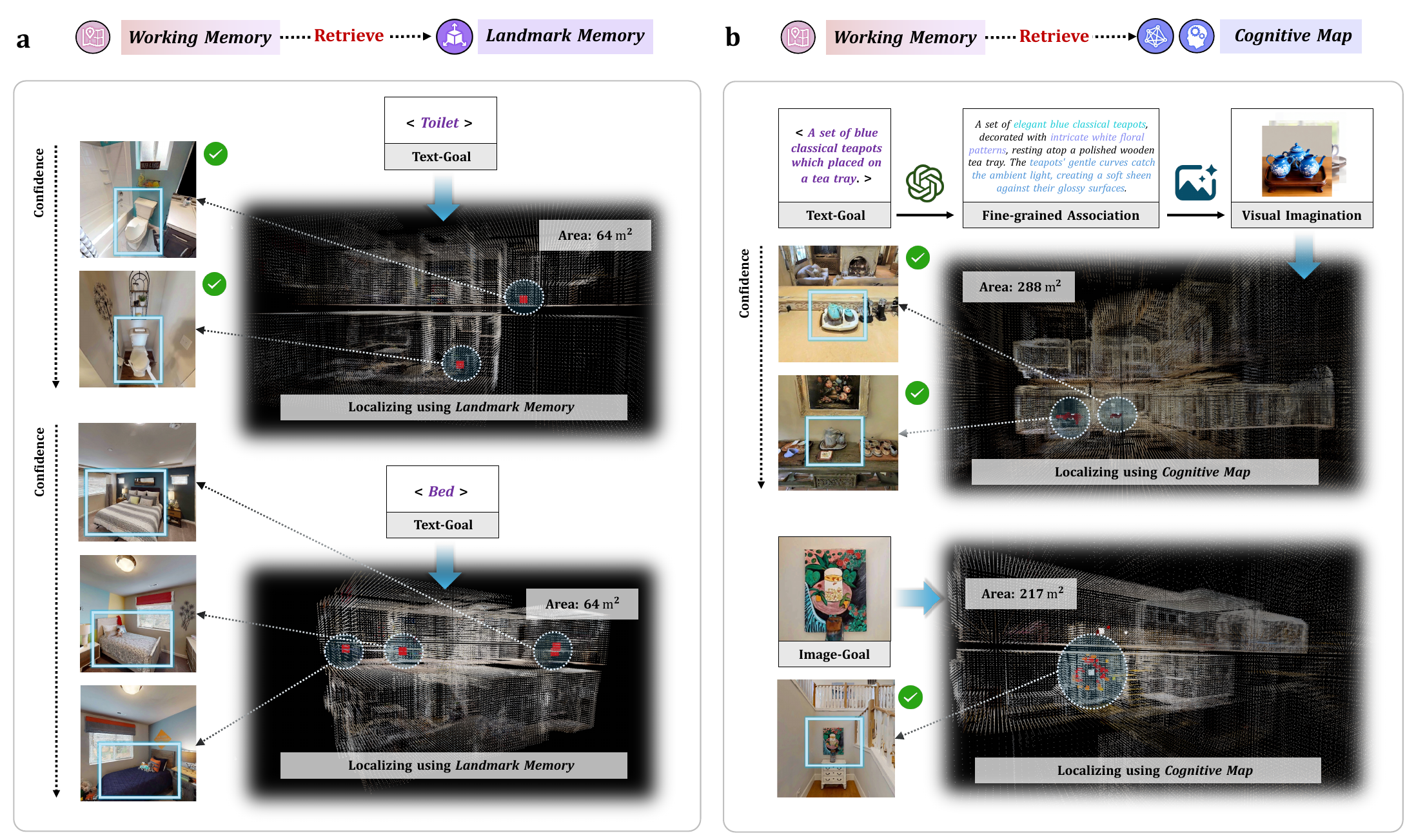}
   \caption{
   \textbf{Precise localization via hierarchical retrieval in working memory.}
   \textbf{a}, For simple category-level goals, working memory prioritizes retrieval from the landmark memory module for rapid matching and candidate coordinate generation.
   \textbf{b}, For complex instance-level goals, working memory employs association-enhanced retrieval, converting text instructions into visual features to query the cognitive map module. For image-based goals, the cognitive map is queried directly using extracted visual features.}
   \label{fig:Figure2}
\end{figure*}

We propose a hierarchical retrieval strategy guided by the complexity of the goal. For simple targets, the agent leverages MLLMs~\cite{gpt-4o, yang2023qwenvl} in a text-only form to reason over semantic associations and contextual cues within the \textbf{landmark memory} (Fig.~\ref{fig:Figure1}c). 
For more complex or ambiguous instructions, BSC-Nav activates association-enhanced retrieval over the \textbf{cognitive map}. In this process, MLLMs enrich the initial goal descriptions by inferring object-specific attributes and scene-level priors, transforming vague commands into semantically detailed representations. These enriched descriptions are then rendered into imagined visual prototypes using a text-to-image diffusion model~\cite{esser2024scaling}. The generated images are encoded and matched against dense visual features stored in the cognitive map to localize candidate target regions.
This hierarchical retrieval strategy allows BSC-Nav to access complementary forms of spatial memory in response to target complexities (category-level and instance-level) and modalities (text and image), supporting precise localization for universal goal-directed navigation (Fig.~\ref{fig:Figure2}).
%task descriptions are first enriched by MLLMs that infer object-specific attributes and scene-level priors, transforming vague goals into detailed semantic representations. These enriched descriptions are then rendered into imagined visual prototypes via text-to-image diffusion models~\cite{esser2024scaling}, which are matched against dense perceptual features stored on the cognitive map to identify candidate locations.

Working memory retrieval often produces multiple candidate coordinates, each associated with a confidence score and relative distance. In scenarios with multiple valid targets (e.g., category-level navigation), confidence scores alone may not reliably indicate correctness. Rather than greedily selecting the candidate with the highest confidence,
%Instead of committing to the prediction of the highest confidence, 
BSC-Nav employs a composite ranking strategy that integrates distance- and confidence-based scores to determine an efficient, adaptive exploration sequence.
%ranks candidates using combined distance- and confidence-based scores, thus determining an efficient and adaptive exploration sequence.
Low-level movement policies are generated using heuristic planning along this sequence. During execution, MLLMs continuously parse spatial-semantic context from observations, verify landmark proximity, and confirm goal arrival, enabling affordance-aware action generation. Upon reaching the appropriate target locations, agents can invoke additional skills, such as grasping objects or answering questions, to accomplish more complex, task-driven objectives (Fig.~\ref{fig:Figure1}c).

\subsection{Universal navigation across modalities and granularities}

Through brain-inspired spatial cognition that incorporates the three types of spatial knowledge, BSC-Nav demonstrates outstanding generalization capabilities in navigation tasks, significantly outperforming recent strong baselines in both success rate and efficiency.

We conduct systematic evaluations across 8,195 episodes in 62 indoor scenes from physically reconstructed environments (MP3D~\cite{chang2017matterport3d} and HM3D~\cite{ramakrishnan2021habitat} datasets) in the Habitat simulator~\cite{savva2019habitat}, covering four representative navigation tasks: \textbf{Object-Goal Navigation (OGN)}~\cite{batra2020objectnav, chaplot2020object, sun2024survey}, \textbf{Open-Vocabulary Object Navigation (OVON)}~\cite{yokoyama2024hm3d}, \textbf{Text-Instance Navigation (TIN)}~\cite{sun2024prioritized}, and \textbf{Image-Instance Navigation (IIN)}~\cite{krantz2022instance}. In each episode, agents initialized at random positions execute discrete actions (forward 25 cm, turn left 30°, turn right 30° and stop)~\cite{gervet2023navigating}. Following standard protocols~\cite{gervet2023navigating}, navigation succeeds only when agents execute stop within 1.0 m of goals. Evaluation metrics include the Success Rate (SR)~\cite{anderson2018evaluation} for \textbf{efficacy} and the Success weighted by Path Length (SPL)~\cite{anderson2018evaluation} for \textbf{efficiency} (detailed in Methods). We compare BSC-Nav with end-to-end methods (PixNav~\cite{cai2024bridging}, DAgRL~\cite{yokoyama2024hm3d}, PSL~\cite{sun2024prioritized}) and modular methods that instantiate only analogous landmark memory (VLFM~\cite{yokoyama2024vlfm}, MOD-IIN~\cite{krantz2023navigating}, UniGoal~\cite{yin2025unigoal}, GOAT~\cite{GOAT2023}).

\begin{figure*}[t]
  \centering
  \includegraphics[width=0.98\linewidth]{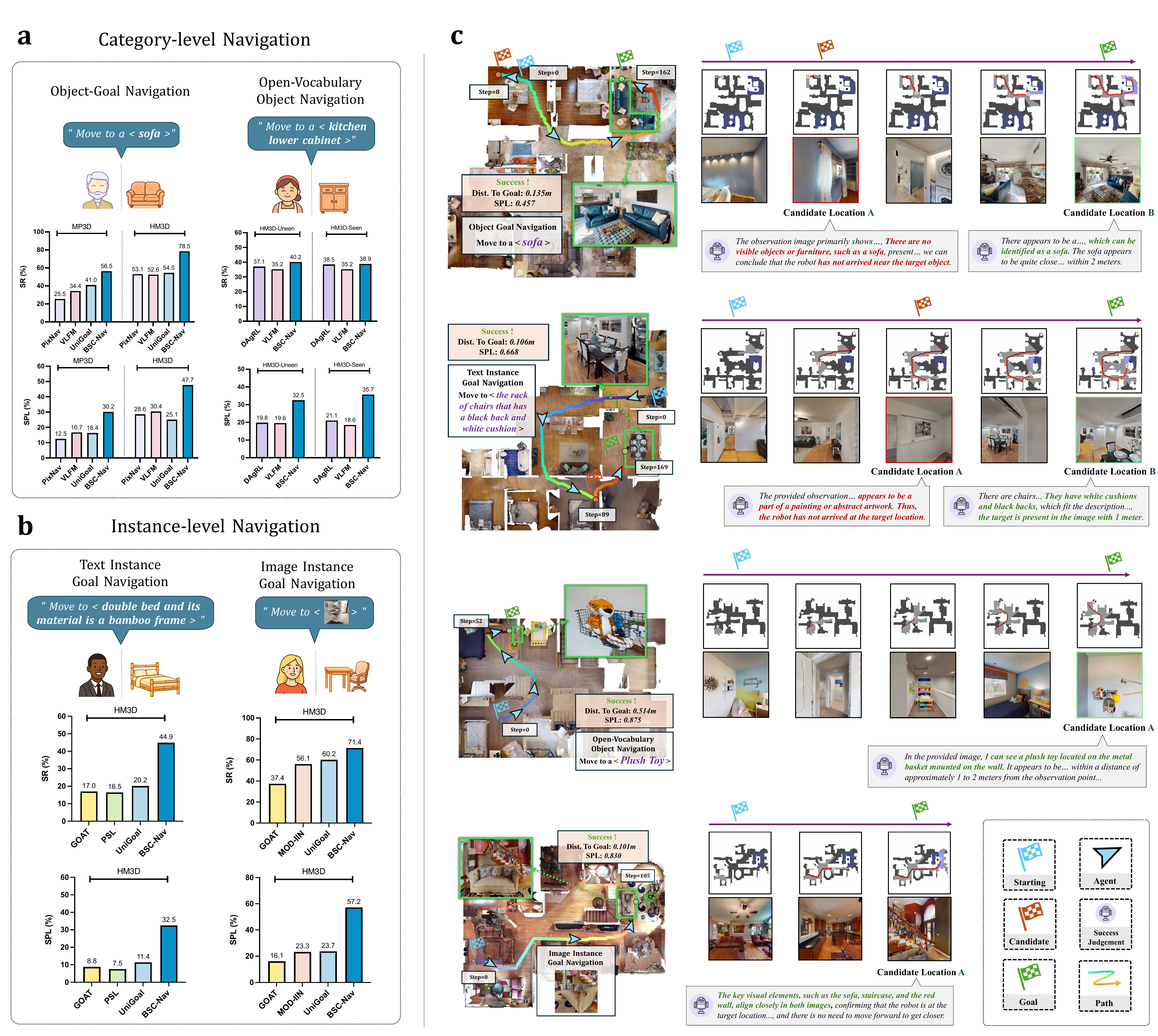}
   \caption{\textbf{Goal-directed multi-modal navigation.}
   %\textbf{a-b}. Experimental results on four foundational navigation tasks. 
   \textbf{a}, Category-level navigation tasks, including object-goal navigation and open-vocabulary object navigation.
   \textbf{b}, Instance-level navigation tasks, including text-instance navigation and image-instance navigation.
   %From left to right, we report results on category-level navigation, including evaluation on object-goal navigation and open-vocabulary object navigation benchmarks (\textbf{a}), and results on instance-level navigation tasks, including text-instance navigation and image instance navigation benchmarks (\textbf{b}). 
   \textbf{c}, Visualization of BSC-Nav navigation trajectories, including the agent's egocentric observations and target verification via MLLM interaction.
   }
   \label{fig:Figure3}
\end{figure*}

As shown in Fig.~\ref{fig:Figure3}a, BSC-Nav demonstrates superior spatial generalization in category-level navigation (OGN and OVON). For common targets in OGN tasks, BSC-Nav achieves 78.5\% SR on HM3D (6 categories) and 56.5\% on MP3D (20 categories with larger areas), surpassing the state-of-the-art method UniGoal by 24.0\% and 15.5\%, respectively. Unlike UniGoal, which organizes only landmark memory with abstract goals and scene graphs, BSC-Nav leverages structured spatial memory and thus obtains significant performance gains.
In the more challenging OVON tasks involving 79 everyday objects (e.g., ``kitchen lower cabinet''), BSC-Nav maintains 40.2\% and 38.9\% SR on MP3D's unseen and seen validation sets, outperforming the supervised method DAgRL in zero-shot settings.
The results of instance-level navigation (TIN and IIN) further validate BSC-Nav's robust multi-modal generalization (Fig.~\ref{fig:Figure3}b). In TIN tasks, it nearly doubles the SR compared to UniGoal and VLFM. In IIN tasks, direct matching of target images with cognitive map features yields 71.4\% SR, outperforming UniGoal by 11.4\%.
%These results demonstrate the critical role of structured spatial memory in enabling precise localization and navigation to multi-modal targets.

Notably, BSC-Nav achieves significantly higher navigation efficiency across all tasks compared to baseline methods, as measured by SPL scores. This improvement stems from our distance- and confidence-based scoring strategy within the working memory module, which enables the agent to plan efficient exploration sequences by examining only a small number of candidate locations (Supplementary Fig.~\ref{fig:SuppFigure2}). 
Representative visualization examples are presented in Fig.~\ref{fig:Figure3}c, including top-to-down trajectories, first-person observations, and MLLM-assisted target verification. We provide additional examples in Supplementary Video~1 and benchmark results in Supplementary Fig.~\ref{fig:SuppFigure2}.

\subsection{Higher-level spatially-aware skills}

By integrating structured spatial memory with MLLMs' robust visual grounding and high-level planning capabilities, BSC-Nav demonstrates strong performance on higher-order spatial tasks, such as long-horizon navigation and spatial reasoning based on complex linguistic instructions.

We evaluate BSC-Nav on a representative task termed \textbf{Long-horizon Instruction-based Navigation (LIN)}~\cite{mattersim, krantz_vlnce_2020}, which requires agents to understand and execute complex instructions containing multiple intermediate goals and spatial constraints. For example, an instruction such as ``Go through the glass door, pass between the sofa and the coffee table, walk to the refrigerator, then turn right and stop at the staircase entrance'' demands nuanced vision-language reasoning and spatial understanding.
To address this, we equip BSC-Nav with GPT-o3~\cite{gpt-o3}, a powerful MLLM with strong reasoning capabilities. 
GPT-o3 decomposes complex instructions into spatially-grounded subgoals (termed \textit{waypoints}, such as ``glass door'', ``refrigerator'', and ``staircase entrance'') by language instruction and initial visual observation of the agent. This hierarchical planning strategy transforms instruction-following into sequential goal-directed navigation steps, enabling BSC-Nav to reliably reach each waypoint and ultimately the final destination.

\begin{figure*}[t]
  \centering
  \includegraphics[width=0.98\linewidth]{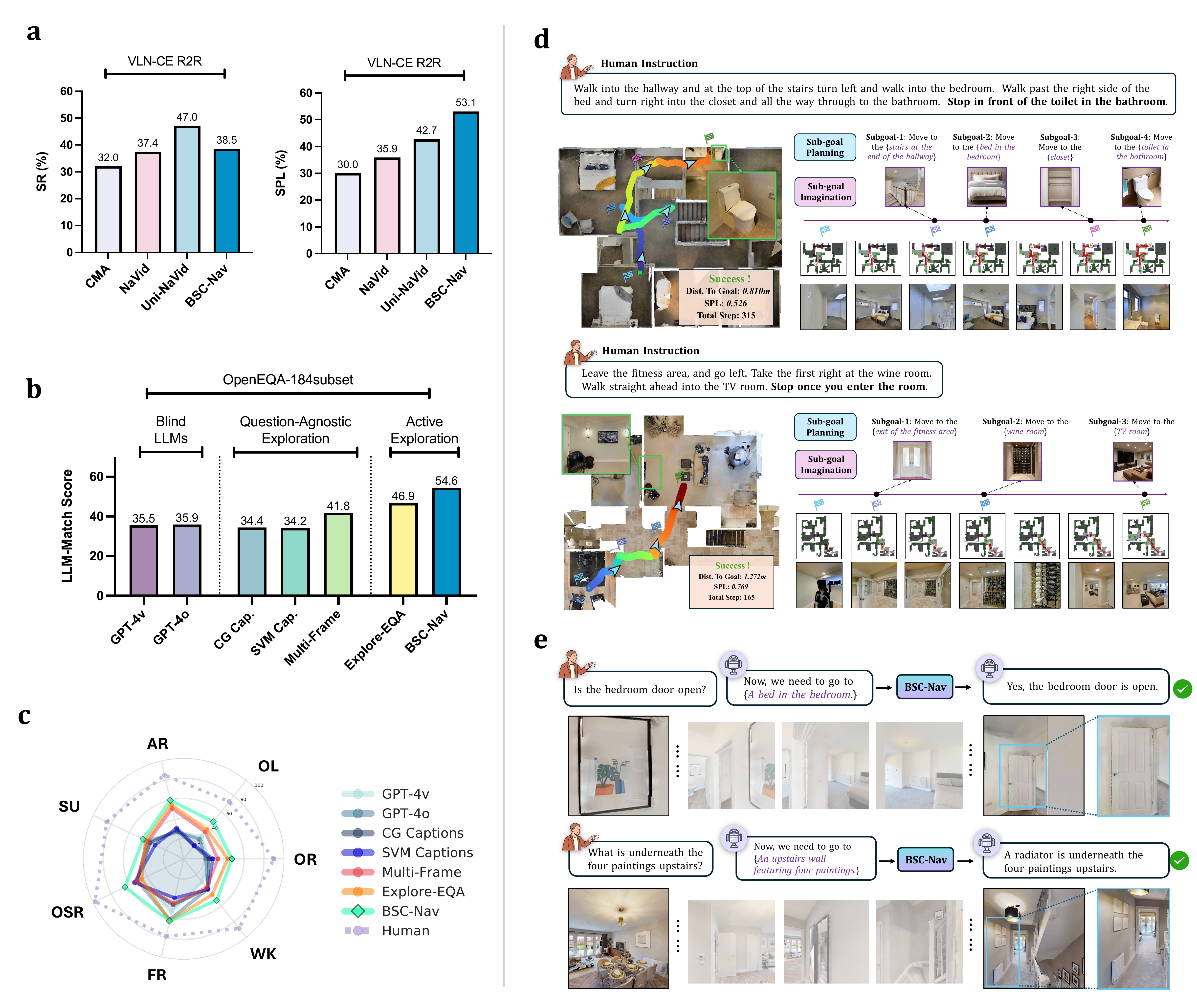}
   \caption{
\textbf{Higher-level spatially-aware skills enabled by BSC-Nav.}
   \textbf{a}, Comparison of BSC-Nav and baseline methods on long-horizon instruction-following tasks using the VLN-CE R2R benchmark.
   %Quantitative results comparing BSC-Nav with different methods on human instruction navigation tasks using the VLN-CE R2R benchmark. 
   \textbf{b}, Comparison of BSC-Nav and baseline methods on active embodied question answering tasks using the A-EQA benchmark. 
   \textbf{c}, Category-wise performance breakdown across seven question types for BSC-Nav, baselines, and humans in embodied question answering tasks.
   \textbf{d}, \textbf{e}, Representative examples of BSC-Nav's navigation trajectories in human instruction navigation (\textbf{d}) and embodied question answering (\textbf{e}) tasks.}
   \label{fig:Figure4}
\end{figure*}

We evaluate BSC-Nav on the VLN-CE Room-to-Room (R2R) benchmark~\cite{krantz_vlnce_2020}, which includes 1,000 human-annotated long-horizon instructions from MP3D. As shown in Fig.~\ref{fig:Figure4}a, BSC-Nav achieves 38.5\% SR in zero-shot settings, only 8.5\% below Uni-Navid~\cite{zhang2024uni}, the state-of-the-art Vision-Language-Action (VLA) model trained with extensive task-specific supervision. Notably, BSC-Nav attains 53.1\% SPL for navigation efficiency, significantly outperforming all baseline methods. 
These results underscore the strength of combining structured spatial memory with foundation models for generalizing to complex, long-horizon spatial tasks, without requiring instruction-level supervised training. Representative navigation trajectories are visualized in Fig.~\ref{fig:Figure4}d, including simplified waypoint descriptions and visual prototypes generated by working memory retrieval. Additional examples are provided in Supplementary Video~2.

The value of spatial cognition extends beyond navigation to spatial reasoning and scene understanding.
Here we consider a representative task termed \textbf{Active Embodied Question Answering (A-EQA)}~\cite{das2018embodied, wijmans2019embodied, majumdar2024openeqa}, which requires agents to answer spatially grounded questions through active exploration of the environment. Unlike static visual question answering, A-EQA demands spatial understanding, exploratory planning, and dynamic observation synthesis, all well aligned with BSC-Nav's strengths. To address each question, BSC-Nav first parses target waypoints (instances or regions relevant to the query) and then conducts goal-directed exploration. Upon reaching the appropriate locations, GPT-4o~\cite{gpt-4o} integrates local observations with the original question to generate spatially grounded answers.
%original questions, generating informed answers grounded in spatial context.

We evaluate BSC-Nav on the A-EQA subset of the OpenEQA benchmark~\cite{majumdar2024openeqa}, which includes 184 questions that span seven categories: Object Recognition (OR), Object Localization (OL), Attribute Recognition (AR), Spatial Understanding (SU), Object State Recognition (OSR), Functional Reasoning (FR), and World Knowledge (WK). 
We compare BSC-Nav with three representative baselines: a blind LLM that answers without visual observations; a question-agnostic frontier exploration strategy~\cite{majumdar2024openeqa}; and Explore-EQA~\cite{ren2024explore}, an active exploration method without structured spatial memory.
The overall performance is measured by LLM-Match~\cite{majumdar2024openeqa}, which computes semantic similarity between the generated and reference answers using LLMs (detailed in Methods). As shown in Fig.~\ref{fig:Figure4}b, BSC-Nav achieves the highest LLM-Match score of 54.6, significantly outperforming all baselines. 
The category-wise breakdown in Fig.~\ref{fig:Figure4}c highlights especially strong improvements in tasks that require fine-grained spatial localization (OL and OSR) for rapid and accurate exploration. 
Despite these advances, a performance gap remains compared to that in humans, underscoring the ongoing challenges of embodied spatial cognition. Representative examples in Fig.~\ref{fig:Figure4}e illustrate how BSC-Nav combines goal-directed exploration with structured spatial memory to address diverse spatial reasoning problems.

%\clearpage
\begin{figure*}[t]
  \vspace{-0.6cm}
  \centering
  \includegraphics[width=0.80\linewidth]{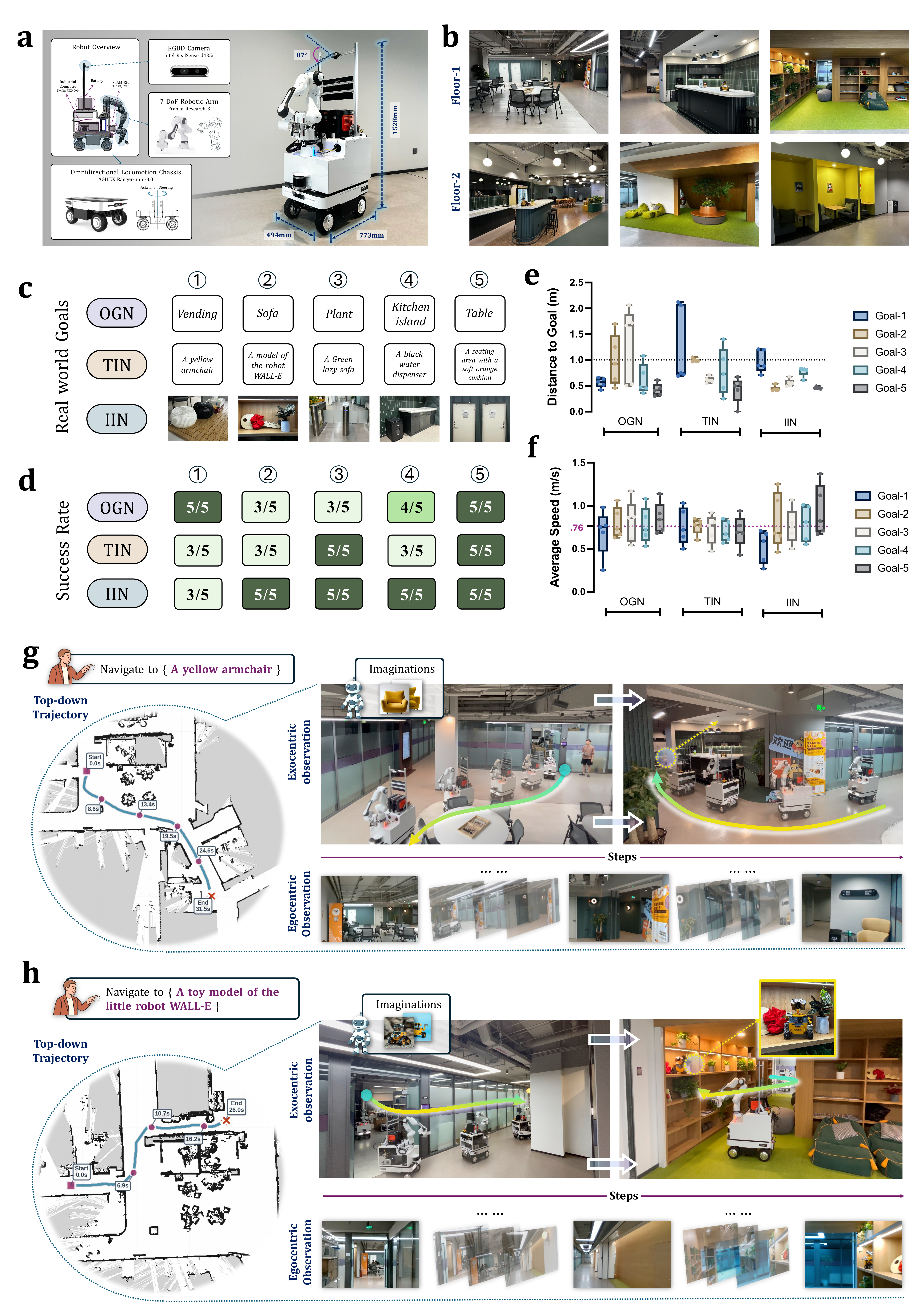}
   \caption{\textbf{Universal navigation capabilities in real-world scenarios.}
   \textbf{a}, The custom-built robotic platform integrates perception, navigation, and manipulation capabilities.
   \textbf{b}, The indoor experimental environment spans 200 m$^2$ and includes diverse functional zones (e.g., office, lounge, reception, and kitchen).
   \textbf{c-f}, Real-world navigation performance across object-goal, text-instance, and image-instance tasks with 15 distinct targets (\textbf{c}). 
   For each target, we report SR from 5 trials with randomized starting positions (\textbf{d}), final distance to goal (\textbf{e}), and mean navigation velocity (\textbf{f}). 
   \textbf{g}, \textbf{h}, Representative examples of real-world navigation. Each shows a top-down trajectory with timestamps (left) and corresponding egocentric/allocentric views (right).}
   \label{fig:Figure5}
    \vspace{-0.2cm}
\end{figure*}
%\clearpage

\begin{figure*}[th]
  \vspace{-0.6cm}
  \centering
  \includegraphics[width=0.80\linewidth]{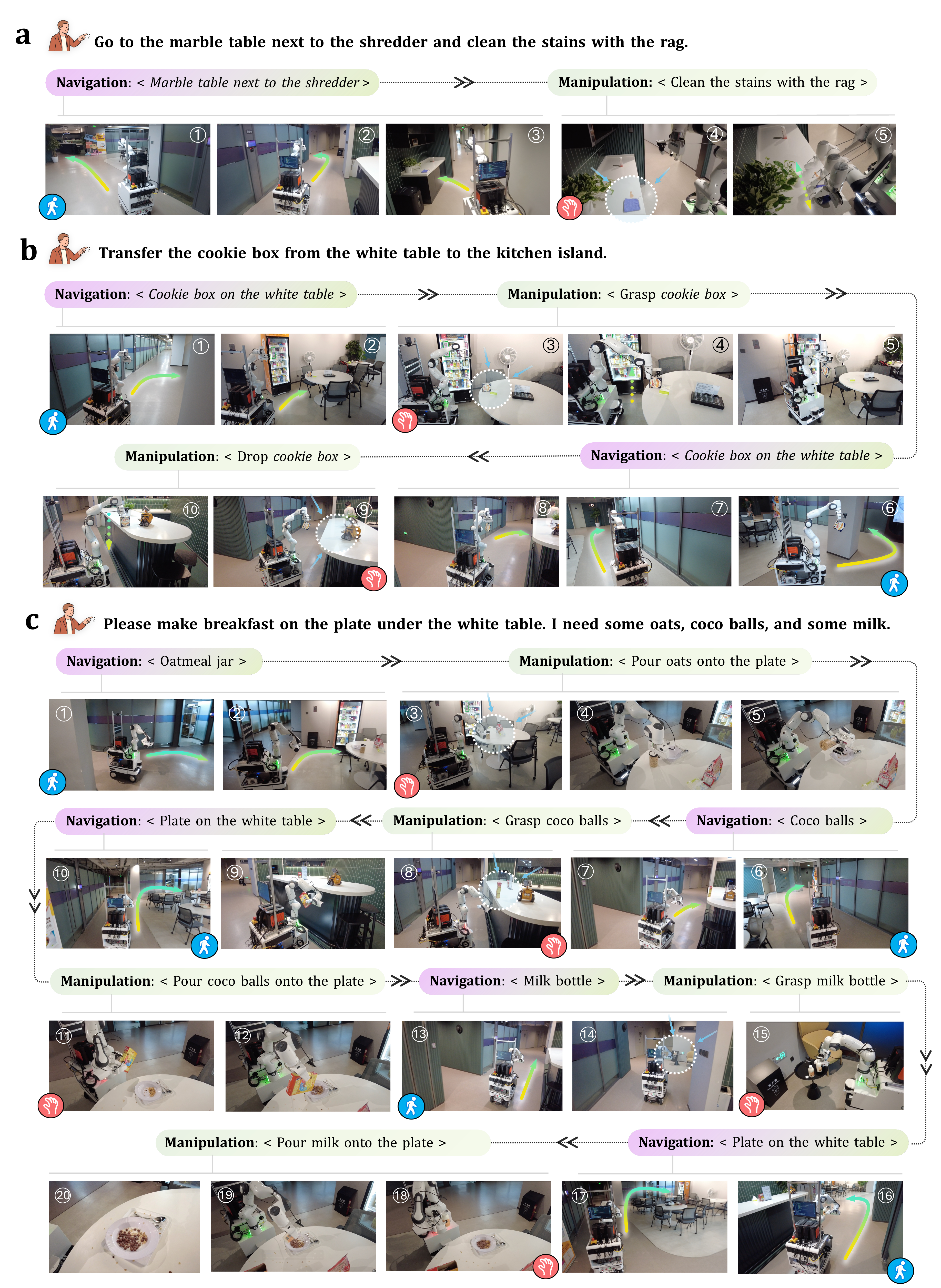}
   \caption{\textbf{Real-world mobile manipulation.} 
   The coordination between spatial memory-driven navigation and manipulation primitives enables the execution of long-horizon tasks specified through human instructions.
   \textbf{a}, Single-waypoint task, cleaning stains on a marble table next to a shredder. 
   \textbf{b}, Object transfer task, relocating a cookie box from table to the kitchen island, requiring navigation between manipulation actions. 
   \textbf{c}, Complex multi-step task, preparing breakfast by sequentially navigating to and manipulating three open-vocabulary targets (oatmeal jar, coco balls, and milk bottle) to assemble ingredients on a plate. 
   Full demonstration are provided in Supplementary video 8-10.}
   \label{fig:Figure6}
   \vspace{-0.2cm}
\end{figure*}
%\clearpage

\subsection{Real-world navigation and mobile manipulation}

To demonstrate BSC-Nav's generalization beyond simulation and its broad applicability in downstream tasks, we deploy it on a custom-built mobile robotic platform within physical indoor environments. This evaluation illustrates how structured spatial memory enables reliable long-range navigation and integrated manipulation in the real physical world.
The mobile platform (Fig.~\ref{fig:Figure5}a) features a compact mechanical design equipped with high-precision motion control, as well as modular sensory and actuation capabilities. We conduct a total of 75 navigation episodes in a two-story indoor space (around 200 m$^2$, Fig.~\ref{fig:Figure5}b), covering three types of goal-directed tasks: \textbf{Object-Goal Navigation (OGN)}, \textbf{Text-Instance Navigation (TIN)}, and \textbf{Image-Instance Navigation (IIN)}, with 5 distinct targets per task (Fig.~\ref{fig:Figure5}c). Each target is tested over 5 trials from randomly sampled starting positions, with an average path length of 23.4 m.

BSC-Nav demonstrates robust spatial generalization in all real-world tasks. As shown in Fig.~\ref{fig:Figure5}d, it achieves at least 3 out of 5 successful trials per target (SR is defined as reaching within 1.0 m), with IIN performing best, reaching 100\% SR on 4 out of 5 targets. Both OGN and TIN reach 100\% SR on 2 targets and exceeds 66.7\% SR across all cases. Even in failure cases, BSC-Nav reliably localizes to semantically plausible regions, demonstrating strong spatial awareness.
The Distance-to-Goal (DtG) distribution in Fig.~\ref{fig:Figure5}e further confirms the precise stop location, with the final DtG below 2.5 m in all cases and tight clustering between targets, particularly for IIN. Navigation efficiency is reflected in a mean velocity of 0.76 m/s with low variance (Fig.~\ref{fig:Figure5}f), indicating stable and efficient movement. Representative navigation trajectories are visualized in Fig.~\ref{fig:Figure5}g–h, with complete demonstrations available in Supplementary Videos 3–7.

Beyond navigation, BSC-Nav's structured spatial memory supports robust \textbf{mobile manipulation} guided by natural language. Although prior embodied manipulation systems~\cite{kim2024openvla, liu2024rdt, black2024pi_0} have shown promise in static, small-scale environments, they often lack the spatial generalization necessary for large-scale deployment. BSC-Nav overcomes this limitation by seamlessly integrating long-horizon navigation with goal-conditioned manipulation. 
In our demonstrations, natural language instructions are interpreted by GPT-4~\cite{wu2023embodied} to generate waypoint-action sequences. Upon reaching each waypoint, the agent executes a predefined manipulation primitive (such as grasp, place, pour, and so on). Representative examples in Fig.\ref{fig:Figure6} show both single-step and multi-step tasks. Notably, the ``make breakfast'' task (Fig.~\ref{fig:Figure6}c) involves identifying and interacting with three spatially distributed, open-vocabulary objects, emphasizing the critical role of structured spatial memory in long-horizon reasoning and reliable object grounding. Full demonstrations are provided in Supplementary Videos 8–10. 

%Together, these results position BSC-Nav as a general-purpose framework for spatial intelligence, capable of bridging large-scale navigation with manipulation and enabling complex embodied behaviors in the real physical world.

\section{Discussion}\label{sec3}

This work demonstrates that biological principles of spatial cognition can be effectively instantiated in embodied agents. By constructing a structured spatial memory comprising landmarks, route knowledge, and survey knowledge, BSC-Nav advances not only navigation performance but also the emergence of cognitive spatial intelligence, which complements the powerful perceptual and reasoning capabilities of modern MLLMs in realizing general-purpose embodied AI systems.

\paragraph{From reactive behavior to cognitive spatial intelligence}

Conventional embodied agents often rely on reinforcement learning or imitation learning, acquiring task-specific policies through extensive trial-and-error or teleoperated demonstrations. While effective in controlled settings, these methods remain fundamentally reactive, responding to immediate stimuli without persistent or reusable spatial knowledge. BSC-Nav addresses this limitation by integrating structured spatial memory with foundation models, enabling a transition from observation-driven pattern matching to multi-level spatial reasoning. Empirically, BSC-Nav exhibits strong performance across navigation tasks of varying modalities and granularities, especially in open-vocabulary and instance-level settings. For example, it achieves 38.5\% SR on the challenging VLN-CE benchmark in a zero-shot setting, coming within 8.5\% of the leading supervised method, while attaining superior efficiency. These results demonstrate how memory-centric spatial representations enable embodied agents to decouple planning from perception, reuse prior experience across tasks, and translate high-level goals into concrete actions, exemplifying the hallmarks of cognitive spatial intelligence.

%\paragraph{Computational insights into biological spatial cognition}
\paragraph{Computational instantiation of biological spatial cognition}

BSC-Nav provides a computational framework that operationalizes long-standing theories of spatial knowledge in biological systems~\cite{werner1997spatial,siegel1975development}. While neuroscience has proposed that spatial cognition builds upon interconnected representations of landmarks, route knowledge, and survey knowledge, most experimental evidence has been behavioral or correlational. BSC-Nav demonstrates how these elements can be synergistically implemented to support long-horizon spatial understanding, planning, and reasoning. In particular, the cognitive map module employs a surprise-driven update strategy inspired by the free-energy principle~\cite{friston2010free}, aligning with the hypothesis that biological brains refine internal models by minimizing prediction errors~\cite{friston2010free,friston2009free}. Furthermore, recent findings suggest that sequential route learning drives the formation of allocentric cognitive maps in the hippocampus-entorhinal circuits~\cite{hilton2023route}, a process mirrored by BSC-Nav's trajectory voxelization. These parallels reveal a convergent computational basis shared between cognitive science and AI.

\paragraph{Toward an embodied Turing test~\cite{zador2023catalyzing} for spatial cognition}

%As embodied AI systems become increasingly capable, there is a growing need to assess not only their perception and control, but also their spatial understanding. 
As embodied AI systems advance in perceptual and motor capabilities, it becomes increasingly important to evaluate their spatial understanding in real-world contexts. Analogous to the classic Turing test in language, one may envision a benchmark for spatial cognition that evaluates whether embodied agents exhibit human-like competence for goal-directed abstraction, planning, and reasoning grounded in physical space. Our study highlights some foundational dimensions:
(i) real-time construction of reusable spatial representations; 
(ii) abstraction and reasoning from sparse, partial observations; and (iii) translation of high-level goals into actionable spatial plans. 
While BSC-Nav exhibits remarkable progress across these dimensions, performance gaps remain. For instance, BSC-Nav achieves a 54.6 LLM-Match score on A-EQA, significantly outperforming prior methods but still trailing human performance by 27.5\%. This underscores the complexity of human spatial cognition, which integrates commonsense knowledge, causal inference, and abstraction from minimal environmental cues. Future benchmarks may formalize a comprehensive embodied Turing test~\cite{zador2023catalyzing} for spatial intelligence, incorporating challenges such as adaptation to environmental changes, multi-step route narration, and collaborative problem-solving.

\paragraph{Outlook and future directions}

BSC-Nav demonstrates that bio-inspired structured spatial memory can substantially enhance generalization and adaptability in embodied AI systems. Future work may focus on scaling the framework to dynamic and unstructured settings, enhancing memory efficiency for real-time deployment, supporting collaborative multi-agent interactions, and incorporating more available sensory modalities. Beyond navigation, this architecture lays the groundwork for broader cognitive functions through hierarchical memory organization, resembling how biological systems coordinate perception, cognition, and decision-making. As current MLLMs-based embodied agents exhibit limited spatial competence, BSC-Nav underscores the promise of memory-centric design in closing this gap. While the pursuit of AGI remains a long-term objective, such advances offer tangible progress toward more capable, adaptable, and cognitively informed AI in the real physical world.

\section{Methods}\label{sec4} %3000 words

\subsection{The BSC-Nav framework}

This section describes the implementation details of BSC-Nav, including the three synergistic modules (Fig.~\ref{fig:Figure1}b) and the pipeline of constructing and leveraging structured spatial memory (Supplementary Fig.~\ref{fig:SuppFigure1}).

\paragraph{Observation space}
The observation space of BSC-Nav is defined as $\mathcal{O}_t = \{I_t, D_t, P_t\}$, comprising the RGB image $I_t \in \mathbb{R}^{H \times W \times 3}$, the depth image $D_t \in \mathbb{R}^{w \times h}$ and the pose of the agent $P_t = (X_t, Y_t, \phi^a_t) \in \mathbb{R}^3$ at time $t$, where $X_t$ and $Y_t$ denote the agent's 2D position in the world coordinate system (defaulting to the Cartesian coordinate centered at the SLAM~\cite{zou2021comparative} initialization point), $\phi_t$ is the angle of yaw, and $H$ and $W$ are image width and height, respectively. For simplicity, we omit the time subscript $t$ in subsequent descriptions. RGB images serve as input for both salient object detection and visual feature extraction, while depth images and pose information enable projection from pixel coordinates to the world coordinate system.

\paragraph{Landmark memory}
During active exploration, an observation $\mathcal{O}$ is processed by two branches in parallel to construct structured spatial memory (Fig.~\ref{fig:Figure2}a). For landmark memory, we instantiate it as a list of 4-tuples as follows:
\begin{equation}
\mathcal{M}_{\text{landmark}} = \{L_k\}^N_{k=1}, \quad  \enspace L_k = \{c_k, \theta_k, \rho_k, \mathcal{T}_k\},
\end{equation}
where $\theta_k = (X_k^w, Y_k^w, Z_k^w) \in \mathbb{R}^3$ represents the 3D coordinates of the center of the $k$-th instance in the world coordinate system, $c_k \in \mathcal{C}$ denotes the open-vocabulary category of the predefined category set $\mathcal{C}$, $\rho_k \in [0,1]$ indicates detection confidence, and $\mathcal{T}_k$ is the description of the instance generated by GPT-4o~\cite{gpt-4o} based on observations, which encompasses texture, shape, and spatial contextual semantics. 

To obtain the world coordinates $\theta_k$, we perform a series of coordinate transformations. Given a detected object with the bounding box center at pixel coordinates $(u_k, v_k)$ and the corresponding depth value $d_k$, we first compute the 3D point in the camera coordinate system, which is defined with its origin at the camera's optical center, the xy-plane parallel to the image plane, and the z-axis aligned with the camera's optical axis. The point in the camera coordinate system can
be calculated through inverse perspective projection:
\begin{equation}
\mathbf{p}_k^{\text{cam}} = d_k \cdot \mathbf{K}^{-1} \begin{bmatrix} u_k \\ v_k \\ 1 \end{bmatrix} = d_k \begin{bmatrix} (u_k - c_x)/f_x \\ (v_k - c_y)/f_y \\ 1 \end{bmatrix},
\end{equation}
where $\mathbf{K} \in \mathbb{R}^{3 \times 3}$ is the intrinsic matrix of the camera with focal lengths $(f_x, f_y)$ and principal point $(c_x, c_y)$. Subsequently, we transform this point into the world coordinate system. Given the robot pose $P_t = (X_t, Y_t, \phi_t)$ from the observation space at time $t$, we construct the base-to-world transformation matrix:
\begin{equation}
\mathbf{T}_{\text{world}}^{\text{base}} = \begin{bmatrix}
\cos\phi_t & -\sin\phi_t & 0 & X_t \\
\sin\phi_t & \cos\phi_t & 0 & Y_t \\
0 & 0 & 1 & Z_{\text{base}} \\
0 & 0 & 0 & 1
\end{bmatrix} \in SE(3),
\end{equation}
where $Z_{\text{base}}$ represents the robot base height in the world frame. The point is then transformed through cascaded homogeneous transformation:
\begin{equation}
\begin{bmatrix} \mathbf{p}_k^{\text{world}} \\ 1 \end{bmatrix} = \mathbf{T}_{\text{world}}^{\text{base}} \mathbf{T}_{\text{base}}^{\text{cam}} \begin{bmatrix} \mathbf{p}_k^{cam} \\ 1 \end{bmatrix},
\end{equation}
where $\mathbf{T}_{\text{base}}^{\text{cam}} \in SE(3)$ represents the fixed rigid transformation from camera to robot base frame. The final world coordinates are extracted as $\theta_k = \mathbf{p}_k^{\text{world}} = (X_k^w, Y_k^w, Z_k^w)^{\top}$. In practice, we employ the open-vocabulary object detector YOLO-World~\cite{cheng2024yolo} for the perception of salient instances. We predefine several common object categories as the landmark category set $\mathcal{C} = \{\text{``sofa''}, \text{``sink''}, \text{``bed''}, \ldots\}$, with a detection confidence threshold to exclude semantically ambiguous and non-salient instances.

To prevent duplicate memorization of the same instance, we perform overlap detection for each newly detected landmark $L_{N+1}$. Define the spatial overlap set as: $\mathcal{U} = \{L_j \in \mathcal{M}_{\text{landmark}} : \|\theta_{N+1} - \theta_j\|_2 < \delta_{\text{overlap}} \wedge c_{N+1} = c_j\}$. If $\mathcal{U} \neq \emptyset$, we perform the memory fusion of existing landmarks: %memory fusion is performed as follow to to generate the fused landmark:
\begin{equation}
L_{\text{fused}} = \begin{cases}
c_{\text{fused}} = c_{N+1} \\
\theta_{\text{fused}} = \frac{\rho_{N+1} \cdot \theta_{N+1} + \sum_{j \in \mathcal{U}} \rho_j \cdot \theta_j}{\rho_{N+1} + \sum_{j \in \mathcal{U}} \rho_j} \\
\rho_{\text{fused}} = \frac{\rho_{N+1} + \sum_{j \in \mathcal{U}} \rho_j}{\lvert \mathcal{U} \rvert + 1} \\
\mathcal{T}_{\text{fused}} = \mathcal{T}_k, \,\, k = \arg\max_{j \in \{N+1\} \cup \mathcal{U}} \rho_j .
\end{cases}
\end{equation}
After memory fusion, all elements in $\mathcal{U}$ are removed from $\mathcal{M}_{\text{landmark}}$ and $L_{\text{fused}}$ is added. This update ensures that each landmark 4-tuple represents a unique spatial instance, avoiding information redundancy.

\paragraph{Cognitive map}

In parallel with the landmark memory module, the cognitive map module leverages RGB observations from exploration trajectories to continuously project visual cues into voxelized visual-spatial representations, progressively integrating and updating route knowledge beyond previous static and simplified representations (Fig.~\ref{fig:Figure2}a). We define the cognitive map as a discrete voxelized representation:
\begin{equation}
\mathcal{M}_{\text{cog}} = \{\mathcal{F}_\mathbf{v}\}_{\mathbf{v} \in \mathcal{V}}, \quad \text{where} \enspace \mathcal{F}_\mathbf{v} = \{f_b\}_{b=1}^{B}, \quad f_b \in \mathbb{R}^\hbar,
\end{equation}
where $\mathcal{V} \subseteq \mathbb{Z}^3$ denotes the discrete voxel index space, each voxel $\mathbf{v} = (v_x, v_y, v_z)$ maintains a feature buffer $\mathcal{F}_\mathbf{v}$ containing up to $B$ feature vectors and $\hbar$ is the dimension of the visual feature. To achieve fine-grained visual-spatial encoding, we employ DINO-v2~\cite{oquab2023dinov2} (a powerful self-supervised visual encoder) to extract patch-level features from continuous 2D RGB observations. %$I$
Given an RGB image $I \in \mathbb{R}^{H \times W \times 3}$, DINO-v2 produces patch tokens organized as a spatial grid $\mathbf{F}_{\text{patch}} \in \mathbb{R}^{H' \times W' \times D}$, where $H' = H/s$ and $W' = W/s$ with patch stride $s$. These patch-level features are projected from 2D image coordinates to corresponding voxel coordinates through a multi-stage transformation process. Let $(i, j)$ denote the indices in the patch grid where $i \in \{0, 1, \ldots, H'-1\}$ and $j \in \{0, 1, \ldots, W'-1\}$. For each patch $(i, j)$, we first determine its center pixel coordinates in the original image:
\begin{equation}
(u_{ij}, v_{ij}) = (j \cdot s + s/2, i \cdot s + s/2),
\end{equation}
where $s$ is the patch stride. Using the depth value $d_{ij}$ sampled at this location, we compute the corresponding 3D point in the camera coordinate system:
\begin{equation}
\mathbf{p}_{ij}^{\text{cam}} = d_{ij} \cdot \mathbf{K}^{-1} \begin{bmatrix} u_{ij} \\ v_{ij} \\ 1 \end{bmatrix}.
\end{equation}
This point is then transformed into world coordinates using the robot pose $P = (X, Y, \phi)$ from the observation space:
\begin{equation}
\begin{bmatrix} \mathbf{p}_{ij}^{\text{world}} \\ 1 \end{bmatrix} = \mathbf{T}_{\text{world}}^{\text{base}} \mathbf{T}_{\text{base}}^{\text{cam}} \begin{bmatrix} \mathbf{p}_{ij}^{\text{cam}} \\ 1 \end{bmatrix},
\end{equation}
where $\mathbf{T}_{\text{world}}^{\text{base}}$ is constructed from the robot pose as defined in Eq. (7). With $\mathbf{p}_{ij}^{\text{world}} = (X_{ij}^w, Y_{ij}^w, Z_{ij}^w)^{\top}$, we discretize the continuous world coordinates into voxel indices:
\begin{equation}
\mathbf{v}_{ij} = \left( \left\lfloor \frac{X_{ij}^w}{\Delta} + \frac{G}{2} \right\rfloor, \left\lfloor \frac{Y_{ij}^w}{\Delta} + \frac{G}{2} \right\rfloor, \left\lfloor \frac{Z_{ij}^w}{\Delta} \right\rfloor \right),
\end{equation}
where $\Delta$ is the voxel size (spatial resolution) and $G$ is the grid dimension. The offset $G/2$ centers the voxel grid at the world origin. The visual feature $\mathbf{F}_{\text{patch}}[i,j] \in \mathbb{R}^D$ from patch $(i,j)$ is then associated with voxel $\mathbf{v}_{ij}$.

In $\mathcal{M}_{\text{cog}}$, we avoid redundantly storing all visual features from new observations, which would lead to information overload and inefficient retrieval. We also forgo conventional fusion methods, such as grid averaging or distance-based weighting~\cite{jatavallabhula2023conceptfusion, huang2023visual}, which often introduce bias in visual feature representation. Instead, inspired by biological learning and memory, we maintain dynamic buffers for each voxel and introduce a surprise-based update strategy. Neuroscience research suggests that biological brains update their internal models by minimizing the difference between predicted and observed sensory inputs, known as the free-energy principle~\cite{friston2009free, friston2010free}. Similarly, BSC-Nav's cognitive map is updated according to the degree of deviation between new observations and existing memory in a given spatial region, that is, the level of ``surprise''. For each new visual feature $f_{\text{new}}$ projected to voxel $\mathbf{v} \in \mathbb{Z}^3$, we compute its surprise score as
\begin{equation}
\mathcal{S}(f_{\text{new}}, \mathbf{v}) = \frac{1}{\left\vert\mathcal{F}_{\mathcal{N}_n(\mathbf{v})}\right\vert} \sum_{f_b \in \mathcal{F}_{\mathcal{N}_n(\mathbf{v})}} \mathcal{D}(f_{\text{new}}, f_b), 
\end{equation}
where $\mathcal{N}_n(\mathbf{v}) = \{\mathbf{v}' \in \mathbb{Z}^3 : \lVert\mathbf{v} - \mathbf{v}'\rVert_\infty \leq n\}$ denotes the $n$-hop cubic neighborhood around voxel $\mathbf{v}$, $\mathcal{F}_{\mathcal{N}_n(\mathbf{v})} = \bigcup_{\mathbf{v}' \in \mathcal{N}_n(\mathbf{v})} \mathcal{F}_{\mathbf{v}'}$ represents the union of all feature buffers in the neighborhood, and $\mathcal{D}(\cdot, \cdot)$ is the distance metric (e.g., cosine distance). We set a predefined threshold $\tau$ that defaults to 0.5. When $\mathcal{S}(f_{\text{new}}, \mathbf{v}) > \tau$, $f_\text{{new}}$ is added to $\mathcal{F}_\mathbf{v}$.
%When $\mathcal{S}(f_{new}, \mathbf{v}) > \tau$ (where $\tau$ is a predefined threshold, defaulting to 0.5), indicating high surprise, $f_{new}$ is added to buffer $\mathcal{F}_\mathbf{v}$. 
If $\lvert\mathcal{F}_\mathbf{v}\rvert = B$, the feature with the lowest surprise score within the buffer is replaced to maintain memory efficiency. This surprise-based update strategy achieves two key advantages: (i) it enhances the robustness of spatial knowledge by selectively caching diverse features across viewpoints and timepoints in adapting to dynamic environments; and (ii) it maintains memory storage and retrieval efficiency by avoiding redundant encoding of stable environmental elements.

\paragraph{Working memory}

The working memory module is responsible for the hierarchical retrieval and strategic reorganization of spatial knowledge from both the landmark memory $\mathcal{M}_{\text{landmark}}$ and the cognitive map $\mathcal{M}_{\text{cog}}$, enabling BSC-Nav to address navigation tasks of varying modalities and granularities (Fig.~\ref{fig:Figure2}b). Unlike the parallel construction and passive updating of the two memory branches, the working memory is activated only upon receiving navigation instructions. It employs a hierarchical retrieval strategy guided by instruction complexity. For simple and concrete goals, it prioritizes the fast retrieval of landmark memory. For fine-grained or image-based instructions, it further engages the cognitive map for precise visual-spatial localization (Fig.~\ref{fig:Figure2}c).

\noindent \textbf{MLLM-reasoning retrieval for landmark memory.} 
The structured knowledge base of landmark memory, including landmark categories, confidence scores, and contextual descriptions, provides a strong foundation for MLLMs to reason about target locations. Unlike direct rule matching, this approach enables context-aware reasoning and can infer the location of unrecorded targets based on spatial associations. For example, even if a ``toaster'' is not explicitly recorded, the system can infer its location by leveraging co-located landmarks such as ``stove'' and ``kitchen island''. Specifically, we design retrieval prompts (see prompt templates in Supplementary Sec.~\ref{sec.prompt}) that guide a text-only GPT-4 to integrate confidence scores and descriptive semantics to generate a set of candidate coordinates $\{\theta^i_{\text{cand}}\}_{i=1}^K$ from $\mathcal{M}_{\text{landmark}}$.

\noindent \textbf{Association-enhanced retrieval for cognitive map.} 
To bridge the modality gap between textual instructions and visual representations, we perform association-enhanced retrieval over the cognitive map. It first employs text-only GPT-4o to refine goal descriptions, enriching them with detailed texture and spatial context information. Specifically, in LIN tasks, initial visual observations are additionally provided as environmental priors to enhance the accuracy and specificity of refined descriptions. These enriched descriptions are then used to ``imagine'' potential visual appearances of targets through a text-to-image generation model, defaulting to Stable~Diffusion~3.5~\cite{rombach2022high}. This ``imagine-then-localize'' process resembles human pre-navigation thinking. The generated imagined image is encoded by DINO-v2 to extract patch-level features $\{f_i\}_{i=1}^Q$, followed by center-distance weighted pooling to obtain the instance visual representation:
\begin{equation}
f_{\text{target}} = \frac{\sum_{i=1}^{N} w_i \cdot p_i}{\sum_{i=1}^{N} w_i}, \quad \text{where} \enspace w_i = \exp(-\alpha \cdot ||(x_i, y_i) - (x_c, y_c)||_2),
\end{equation}
where $(x_i, y_i)$ denotes the spatial coordinates of the $i$-th patch, $(x_c, y_c)$ is the image center, and $\alpha$ is the temperature parameter. This weighting strategy suppresses background interference and enhances central features of the target instance. As an intermediate query, the pooled visual features are further matched against features stored in the cognitive map, returning the top-$K$ voxel coordinate set with maximum cosine similarity. We perform similarity-weighted DBSCAN clustering~\cite{kriegel2011density} on the voxel coordinate set to obtain the grid coordinates of cluster centers as the final spatial coordinate candidates. These grid coordinates need to be further back-projected to the world coordinate system, yielding $\{\theta^i_{\text{cand}}\}_{i=1}^Q$ for subsequent low-level planning.

\noindent \textbf{Exploration sequence planning.} We design a composite scoring function that integrates target existence probability and spatial distance to prioritize the set of candidate spatial coordinates. Specifically, for candidates retrieved from the landmark memory, we use detection confidence as the existence probability. For candidates retrieved from the cognitive map, we employ the cosine similarity of visual features. The priority scoring function is defined as:
\begin{equation}
H_i = \lambda \cdot p_i + (1-\lambda) \cdot (1 - \frac{d_i}{d_{\max}}),
\end{equation}
where $p_i$ represents the existence probability (confidence or similarity) of the $i$-th candidate, $d_i$ is the Euclidean distance between the candidate and the starting point, $d_{\max} = \max_j d_j$ normalizes distance across candidates. The weighting hyperparameter $\lambda$ balances existence probability and exploration efficiency, defaulting to $\lambda = 0.5$ as equal importance to both components.
%assigning equal importance to the efficiency of exploration and the probability of the target.

\paragraph{Low-level navigation policy generation}

Low-level navigation policies are typically derived from deterministic path planning conditioned on environmental constraints and high-level spatial goals~\cite{gervet2023navigating}. BSC-Nav employs a hierarchical navigation strategy: the exploration sequence of each candidate coordinate serves as a high-level policy, while heuristic algorithms are used to generate actionable low-level policies toward each candidate coordinate. In simulation environments, we employ the greedy shortest path algorithm provided by the Habitat simulator~\cite{savva2019habitat}, which operates directly on 3D mesh representations of the scene to compute optimal action sequences within a discrete action space.
For real-world deployment, we implement a two-tier planning architecture. Global planning is carried out using the A* algorithm~\cite{hart1968formal} on occupancy grid maps constructed via LiDAR-based SLAM, yielding globally optimal paths. Local planning leverages the Timed Elastic Band (TEB) algorithm~\cite{rosmann2012trajectory, rosmann2013efficient}, which dynamically adjusts trajectories to avoid obstacles while maintaining efficiency. The TEB planner outputs continuous velocity commands to control the robot chassis, ensuring smooth and precise motion execution.

\paragraph{Goal verification and affordance}

Upon arriving at a candidate coordinate, BSC-Nav verifies whether the navigation target is present at the location. The robot first performs a 360° rotational scan to capture a sequence of RGB images. Cosine similarities are then computed between the CLIP visual embeddings of these images and the goal's text or visual embeddings to identify the viewing angle with the highest semantic alignment. The selected image is subsequently fed into GPT-4o for precise target verification. Beyond confirming target presence, GPT-4o is also prompted to generate a sequence of affordance-based actions to guide the robot in fine-tuning its pose, adjusting its relative position, and orientation for optimal proximity and visibility. This ensures favorable initial conditions for downstream manipulation tasks.

\subsection{Experimental details}

\paragraph{Simulator and datasets}

The simulation experiments are conducted under the Habitat 3.0 platform~\cite{puig2023habitat}, a commonly used simulation framework for embodied AI and human–robot interaction in domestic environments. 
We select this platform for its strong support for large-scale embodied navigation, including (i) habitat-sim, a high-performance simulator offering photorealistic rendering and physics-based interactions for mainstream indoor datasets, and (ii) habitat-lab, a modular benchmarking suite supporting a variety of navigation tasks with standardized pipelines and metrics.
Here we evaluate BSC-Nav and baseline methods on four foundational navigation tasks:
\begin{enumerate}
\item \textbf{Object-Goal Navigation (OGN).} %As featured in Meta's long-standing ObjectNav Challenge~\cite{habitatchallenge2023}, 
This task~\cite{habitatchallenge2023} comprises two benchmarks, including (i) 2,195 episodes across 34 HM3D scenes involving 6 object categories, and (ii) 2,000 episodes across 10 MP3D scenes with 20 object categories.

\item \textbf{Open-Vocabulary Object Navigation (OVON).} 
This task~\cite{yokoyama2024hm3d} extends to 79 open-vocabulary categories across 10 MP3D scenes to address the limited object categories in object-goal navigation.
%To address the limited object categories in object-goal navigation, this task extends to 79 open-vocabulary categories across 10 MP3D scenes. 
We sampled 1,000 episodes each from the validation-seen and validation-unseen splits. The seen split includes categories present in training (though not exact instances), while the unseen split comprises entirely novel and semantically dissimilar categories.

\item \textbf{Text-Instance Navigation (TIN).} This task~\cite{sun2024prioritized} provides natural language descriptions for 795 instances across 36 HM3D scenes. Descriptions encompass both intrinsic attributes (inherent object properties including shape, color, and material) and extrinsic attributes (surrounding environmental contexts), annotated with a strong MLLM termed CogVLM~\cite{wang2024cogvlm}. We evaluate all 1,000 test episodes.

\item \textbf{Image-Instance Navigation (IIN).} This task~\cite{krantz2022instance} employs single-view rendered images as navigation targets for instances across 34 HM3D scenes. We sampled 1,000 episodes from the validation split.
\end{enumerate}

Beyond these foundational navigation tasks, we further evaluate BSC-Nav and baseline methods on higher-level spatially-aware skills:
\begin{enumerate}
\item \textbf{Long-horizon Instruction-based Navigation (LIN).} We employ the VLN-CE Room-to-Room (R2R) benchmark~\cite{krantz_vlnce_2020}, which provides 1,000 navigation episodes in Habitat-lab based on human-annotated long-horizon instructions across 11 MP3D scenes.
\item \textbf{Active Embodied Question Answering (A-EQA).} We develop a custom evaluation pipeline within Habitat-lab using OpenEQA~\cite{majumdar2024openeqa}. Agents are initialized at the first frame of each recorded exploration trajectory and are allowed to actively explore the environment to answer a given spatially grounded question. We evaluate the 184 test queries that span seven task categories.
\end{enumerate}

\paragraph{Evaluation metrics}
We adopt two-dimensional metrics~\cite{anderson2018evaluation} to quantify navigation performance. \textbf{Success Rate (SR)} measures the proportion of successful navigation episodes relative to the total test episodes, assessing the agent's capability to accurately navigate target object instances. Notably, in OGN, any instance of the target category constitutes a valid goal regardless of distance from the starting position, whereas instance-level navigation requires reaching a unique specified instance in the environment. Specifically, an episode is deemed successful if the Euclidean distance between the agent and the target object is within 1.0 m upon executing the ``stop'' action. While \textbf{Success weighted by Path Length (SPL)} jointly considers navigation success and path efficiency:
\begin{equation} 
\text{SPL} = \frac{1}{N}\sum_{i=1}^{N} S_i \cdot \frac{L_i^*}{\max(L_i, L_i^*)},
\end{equation}
where $N$ denotes the total number of episodes, $S_i$ is the binary success indicator for episode $i$ (1 for success, 0 for failure), $L_i^*$ represents the geodesic shortest path length, and $L_i$ is the actual trajectory length executed by the agent. SPL values range from 0 to 1, with higher values indicating more efficient navigation. While SR quantifies task completion capability, SPL further evaluates the optimality of path planning, collectively providing a comprehensive assessment of navigation performance.

For A-EQA tasks, we adopt the LLM-Match scoring metric proposed in Open-EQA~\cite{majumdar2024openeqa} to assess the correctness of agent responses. This metric accounts for the open-vocabulary nature of the agent responses, serving as a substitute for manual evaluation. Specifically, given a question $\mathcal{Q}_i$, a human annotated response $\mathcal{A}^*_i$, and an agent response $\mathcal{A}_i$, the LLM is prompted to assign a score $ \sigma_i \in \{1, ..., 5\}$, where $\sigma_i=1$ denotes an incorrect answer, $\sigma_i=5$ denotes a correct answer, and intermediate values reflect varying degrees of partial correctness. The overall LLM-based correctness is then computed as follows:
\begin{equation} 
\text{LLM-Match} =\frac{1}{N_Q} \sum_{i}^{N_Q} \frac{\sigma_{i}-1}{4} \times 100 \%,
\end{equation}
where $N_Q$ denotes the total number of questions. Following the Open-EQA protocol, we employ text-only GPT-4 under official prompts to ensure evaluation fairness.

\paragraph{Hardware stack}
We develop an embodied platform for BSC-Nav deployment, comprising five core components: locomotion chassis, industrial computer, SLAM suite, robotic arm, and vision sensors. 

The locomotion system employs the Agilex Ranger-mini 3.0 platform, chosen for its balance between cost and payload capacity. It features Ackermann steering with zero-radius turning capability, enabling agile maneuvering in complex environments. The SLAM suite integrates a 32-beam LiDAR, IMU sensor, and depth compensation camera to support collision detection and real-world localization. For manipulation capabilities, we equipped the platform with a Franka Emika Research 3 robotic arm. Two Intel RealSense D435i cameras serve as primary vision sensors (one mounted at 1.5 m above ground level and another at the manipulator's end effector), providing $848\times480$ RGB-D imagery with an 87° field of view. To mitigate depth-sensing artifacts including holes and edge discontinuities, we apply spatio-temporal and hole-filling filters, constraining the effective sensing range to 0.3–8.0 m. An industrial-grade computer equipped with an NVIDIA RTX 4090 GPU handles all real-time processing, including BSC-Nav, SLAM computation, and manipulator control, ensuring end-to-end system responsiveness.
%The industrial computer, equipped with an NVIDIA RTX 4090 GPU, executes BSC-Nav, SLAM algorithms, and manipulator control in real time, ensuring end-to-end system responsiveness.

\paragraph{Implementation details}

Before task execution, BSC-Nav requires environmental perception to construct preliminary landmark memory and cognitive map representations. For simulation environments, we implement a frontier-based~\cite{gervet2023navigating} autonomous exploration strategy for spatial memory construction: at each time step, the system generates a height map through depth projection to identify boundaries between explored and unexplored regions. On traversable boundaries, the system selects the nearest frontier point to the current position as the next exploration target, utilizing Habitat's scene-mesh-based greedy navigator for low-level action planning. Upon reaching each frontier point, the agent performs a 360-degree rotation to comprehensively perceive the surrounding environment. This exploration process continues until a predetermined iteration limit is reached, which is dynamically adjusted based on scene scale. The iteration count is defined as half of the traversable area.
%(specifically: iteration count = traversable area ($m^2$) / 2).

For real-world deployment, safety constraints necessitate manual teleoperation for pre-collecting environmental observations to construct structured spatial memory. Detailed implementations of these methods are available in our code repository.

During task execution, both landmark memory and cognitive map maintain continuous updates. Parameter configurations are as follows: for landmark memory, the detector confidence threshold is set to $0.55$ with a spatial overlap distance of 1.0 m; for cognitive map, we configure a voxel resolution of $\delta = 0.1$ m$^3$ with grid dimension $G=1000$ and buffer capacity $B=10$ per voxel. For each navigation episode, the system first retrieves candidate target locations through the working memory module, with maximum candidates $K=3$ for landmark memory retrieval and $Q=3$ for cognitive map retrieval. We employ Stable Diffusion 3.5-Medium for cognitive map retrieval, generating three images per batch to mitigate stochasticity in visual imagination. The system subsequently navigates to each candidate location following the planned exploration sequence, performing target verification upon arrival at each candidate. Task success is determined when at least one candidate passes verification. If all candidates fail verification, the task is deemed unsuccessful.

\paragraph{Baselines}

\noindent \textbf{End-to-end navigation methods} leverage deep neural networks to directly map egocentric observations to action sequences, implicitly encoding spatial-geometric priors and semantic knowledge within network parameters. They must simultaneously learn spatial memory tracking and action planning, requiring extensive trajectory data or expert demonstrations for effective training. For example, PixNav~\cite{cai2024bridging}, which performs greedy navigation by predicting optimal actions toward salient pixels in the current view. DAgRL~\cite{yokoyama2024hm3d}, which integrates pre-trained vision–language encoders with historical action embeddings via transformer architectures and applies DAgger-based~\cite{ross2011reduction} online policy optimization.
PSL~\cite{sun2024prioritized}, which uses CLIP to encode both visual observations and textual goals, minimizing semantic discrepancies between them. Although these methods have shown strong performance in simulation environments, they often struggle to generalize to the real world due to visual domain shifts. Recent Vision-Language-Action (VLA) models~\cite{zhang2024navid, zhang2024uninavid, cheng2024navila} address these limitations by scaling model capacity and diversifying training trajectories. Uni-NavId exemplifies this trend, achieving strong performance in parsing complex multi-modal instructions. However, these large-scale models incur significant computational overhead, limiting their action generation frequency, and still lack mechanisms for persistent spatial memory integration.

\textbf{Modular navigation methods} offer a more interpretable and adaptable alternative by constructing explicit spatial representations to support the generation of low-level policies. 
For example, GOAT~\cite{GOAT2023} projects closed set semantic segmentation onto top-down semantic maps and learns policies to predict optimal sub-goals from these maps. 
MOD-IIN~\cite{krantz2023navigating} extends this paradigm to image-goal navigation. 
VLFM~\cite{yokoyama2024vlfm} employs the vision-language model, BLIP-2~\cite{li2023blip}, to construct probabilistic frontier maps indicating target likelihood at candidate viewpoints. 
UniGoal~\cite{yin2025unigoal} unifies object categories, instance images, and text descriptions through abstract graph-based representations to support multi-granular spatial modeling. By decoupling low-level control from perceptual inference, these methods exhibit improved sim-to-real transferability. However, current implementations often suffer from incomplete spatial memory by modeling either landmarks or route knowledge in isolation, limiting their performance in higher-level spatial reasoning.

\backmatter

\section*{Data Availability}
All simulation scene data and evaluation benchmarks used in this study are publicly available through open-source platforms or datasets. The MP3D and HM3D scene data can be accessed through the Habitat-sim repository (\href{https://github.com/facebookresearch/habitat-sim/blob/main/DATASETS.md}{}). Navigation benchmarks including object-goal navigation, image-instance navigation, and VLN-CE R2R datasets are available through Habitat-lab (\href{https://github.com/facebookresearch/habitat-lab/blob/main/DATASETS.md}{}). The open-vocabulary object navigation benchmark and configuration files are from Yokoyama et al.~\cite{yokoyama2024hm3d} (\href{https://github.com/naokiyokoyama/ovon}{}). Text-instance navigation data and configurations are from Sun et al.~\cite{sun2024prioritized} (\href{https://github.com/XinyuSun/PSL-InstanceNav}{}). Active embodied question-answering data are from the OpenEQA dataset by Majumdar et al.~\cite{majumdar2024openeqa} (\href{https://github.com/facebookresearch/open-eqa}{}).

\section*{Code Availability}
The BSC-Nav implementation for both simulation and real-world experiments, along with data analysis and visualization scripts, is publicly available on GitHub (\href{https://github.com/Heathcliff-saku/BSC-Nav}{}). The repository is organized into two branches: the \textit{sim} branch contains simulation implementations, benchmark configurations, and evaluation scripts; the \textit{phy} branch includes low-level control interfaces adapted for our robotic platform and BSC-Nav deployment scripts. Researchers can refer to these implementations to deploy BSC-Nav on custom-built embodied agents. 
% An archived version of the code is also available through Zenodo ().

\section*{Acknowledgments}
This work was supported by NSFC Projects (Nos. 92370124, 92248303, 62276149, 62350080, 62406160), BNRist (BNR2022RC01006).

\section*{Author Contributions Statement}

S.R., L.W. and H.S. conceived the study. S.R., L.W. and S.L. designed the computational model. S.R. and C.K. performed the main experiments, assisted by Q.Z.. S.R. and Q.Z. completed the design of the embodied robot and algorithm deployment. L.W. and S.R. drafted the paper. H.S. and X.W. supervised the project. H.S. offered funding and critically revised the paper. All authors read and approved the final version. 

\section*{Competing Interests Statement}
The authors declare no competing interests.

\clearpage

\bibliography{sn-bibliography}% common bib file


\clearpage

\appendix

\section{Prompt Template}\label{sec.prompt}

BSC-Nav deeply integrates the perceptual and reasoning capabilities of Multi-modal Large Language Models (MLLMs) across multiple critical system components, leveraging their strengths in visual understanding, semantic reasoning, and task planning. To fully harness the potential of MLLMs and ensure their reliability in various tasks, we develop a comprehensive prompt engineering encompassing core functional modules including target verification, instruction decomposition, etc. Below provide core prompt template for each functional module. For specific model versions, and complete prompt implementations, please kindly refer to our code repository.

\definecolor{darkblue}{RGB}{0, 0, 139}
\definecolor{lightblue}{RGB}{173, 216, 230}

\begin{framed}
\noindent \colorbox{lightblue}{\textcolor{darkblue}{\textbf{\textit{Prompt for Landmark Memory Retrieval}}}}
\vspace{0.3cm}

\noindent
Given a textual description of a navigation target and a landmark memory list containing instances of detected objects, each with a label, a concise description, a 3D location (three numerical coordinates), and a confidence score, you need to determine the most suitable memory instance to fulfill the navigation request.

\noindent \textbf{\textit{1. Landmark memory data structure:}} 

\noindent The memory is a list of objects in the environment, where each object is represented as a JSON-like structure of the following form.
\begin{lstlisting}
{
    label: <string>,
    description: <string>,
    loc: [<float>, <float>, <float>],
    confidence: <float>
}
\end{lstlisting}
\textbf{\textit{2. Your task:}}

\noindent\textit{2.1 Understand the target description:} You will be given a textual goal description of a navigation target, such as “A marble island in a kitchen.” Your first step is to interpret this description and deduce which object label or description from the memory best matches it semantically.

\noindent\textit{2.2 Identify the relevant instances:} Once you have determined the best matching label, filter the memory list to only those instances whose label matches (or closely matches) that label.

\noindent\textit{2.3 Evaluate confidence and consolidate duplicates:} Among these filtered instances, consider that multiple memory entries may actually represent the same object, possibly due to partial overlaps or multiple detections. 1) Look at their loc coordinates. If multiple instances with the same label have very close or nearly identical coordinates, treat them as the same object. 2) Determine which set of coordinates (if there are multiple distinct sets) is the most reliable representation of the object. Reliability is judged primarily by the highest confidence value. If multiple instances cluster together with similar locations, select the one with the highest confidence or, if confidence is similar, the one that best aligns with the object as described.

\noindent\textit{2.4 Select the final loc:} After you have grouped instances and decided which group best represents the target object, output the coordinates (loc) of the best match. If multiple objects ($\geq$3 items) match the description equally well, choose the three coordinates (loc) with the highest confidence.

\noindent\textit{2.5 Produce a final answer:} Return the selected location coordinates as the final answer, must be in the format

\noindent\sethlcolor{gray!10}\hl{\{Nav Loc 1: [...], Nav Loc 2: [...], Nav Loc 3: [...]\}} 

\noindent\sethlcolor{gray!10}\hl{\{Nav Loc: Unable to find\}}

\noindent\textbf{\textit{3. Important details}}

\noindent3.1 Always provide reasoning internally (you may do it in hidden scratchpads if available) before giving the final result. The final user-visible answer should be concise and directly address the task.

\noindent3.2 If no objects are found that are semantically relevant to the target description, explicitly indicate that no suitable object was found.

\noindent3.3 Follow these steps for every input you receive. 

\noindent \sethlcolor{yellow!35}\hl{Navigation target: \{text\_prompt\}}

\noindent \sethlcolor{yellow!35}\hl{Memory: \{landmark\_memory\}}

\noindent Now please start thinking one step at a time and then Briefly tell me the target locations.
\end{framed}

\begin{framed}
\noindent \colorbox{lightblue}{\textcolor{darkblue}{\textbf{\textit{Prompt Template for Association-Enhanced Text Generation}}}}
\vspace{0.3cm}

\noindent
You are an expert in generating prompts for text-to-image models. Your task is to enhance a given original goal description, which often only mentions a general category or a simple phrase describing the target object, by incorporating detailed context from four current scene images.

\noindent \textbf{\textit{1. Task Overview:}} 
\noindent Create a more imaginative and contextually enriched description, using elements observed in the scene to create a coherent and vivid visual. This description will guide a text-to-image model to generate an image that aligns with the style and context of the current scene. The target object must remain the primary visual focus.

\noindent \textbf{\textit{2. Suggested Steps:}}

\noindent\textit{2.1 Understand the Environment:} Extract and comprehend details from the provided observation images, such as the overall style, decoration, and elements of the scene. Consider whether this is a modern home, classical residence, exhibition hall, or office.

\noindent\textit{2.2 Expand the Original Description:} Based on the scene analysis, enrich the original description with finer details including materials, colors, textures, placement, and environmental elements.

\noindent\textit{2.3 Maintain Visual Focus:} Ensure additional context or background details do not overshadow the main target object. The primary subject should remain the focal point, using language that emphasizes its prominence.

\noindent \textbf{\textit{3. Guidelines for Creating Enhanced Descriptions:}}
\noindent\textit{3.1 Details:} Include sensory details like colors, textures, lighting, and reflections.

\noindent\textit{3.2 Background Elements:} Add appropriate background elements that complement the scene without detracting from focus.

\noindent\textit{3.3 Focus Phrasing:} Use language that naturally draws attention to the target object (e.g., ``centered'', ``prominently placed'', ``as the focal point'').

\noindent\textit{3.4 Balance:} Strike a balance between richness and simplicity. The target object should always dominate the final image.

\noindent \textbf{\textit{4. Output Requirements:}}

\noindent 4.1 Provide a refined and detailed description in English.

\noindent 4.2 Ensure the enhancement creates a vivid, coherent, and engaging scene.

\noindent 4.3 Avoid overly complex narratives or distracting elements.

\noindent 4.4 Keep the description concise, limiting it to 70 words or less.

\noindent \textbf{\textit{5. Examples:}}

\noindent\textit{Example 1:}
\begin{itemize}
\item Original: A green vase.
\item Enhanced: A vibrant green ceramic vase, with a glossy, smooth surface, placed centrally on a polished wooden table. Soft natural light illuminates the vase from the large window behind it, casting shadows on the table. The surrounding room is decorated in minimalist modern style with neutral tones, ensuring the vase is the central focal point.
\end{itemize}

\noindent\textit{Example 2:}
\begin{itemize}
\item Original: A armchair.
\item Enhanced: A sleek, modern blue armchair, upholstered in soft velvet, positioned prominently in a stylish living room. The chair is placed near a large floor-to-ceiling window, allowing natural light to highlight its deep blue hue. The room features minimalist decor with white walls, light wood flooring, and a few abstract art pieces on the walls.
\end{itemize}

\noindent \sethlcolor{yellow!35}\hl{Original: \{text\_prompt\}}, 

\noindent \sethlcolor{yellow!35}\hl{Observations: \{[img\_list]\}}.

\noindent Please follow the above requirements and examples to enhance this description, think step by step and give your analysis process and the final enhancement description following the format:

\noindent\sethlcolor{gray!10}\hl{analysis process: [your analysis process here]}

\noindent\sethlcolor{gray!10}\hl{enhancement description: [your enhancement description here]}
\end{framed}

\begin{framed}
\noindent \colorbox{lightblue}{\textcolor{darkblue}{\textbf{\textit{Prompt for Goal Verification and Affordance}}}}
\vspace{0.3cm}

\noindent
You will be provided with a navigation observation image from a robot, and a textual description of the navigation goal. Please follow the steps below to determine whether the current navigation task is successful.

\noindent \textbf{\textit{1. Determine Target Presence:}} 

\noindent Analyze the provided images to ascertain whether the navigation goal is present in these images AND close enough (within 2 meters). This means evaluating if the robot has arrived near the target location. Be careful not to misclassify similar categories (e.g. sofas and chairs are easily confused).

\noindent \textbf{\textit{2. Determine whether need to move forward:}}

\noindent If you have found the target according to step 1, you need to further determine whether you need to move forward a small step to get closer to the target object. If need to, answer ``need forward: yes''. If you think you are close enough (within 1m), answer ``need forward: no''.

\noindent \textbf{\textit{3. Output Format:}}

\noindent\textit{3.1 First Line:} \sethlcolor{gray!10}\hl{Success: yes} OR \sethlcolor{gray!10}\hl{Success: no}

\noindent\textit{3.2 Second Line (only when ``success: yes''):} \sethlcolor{gray!10}\hl{need forward: yes} OR \sethlcolor{gray!10}\hl{need forward: no}

\noindent\textit{3.3 Third Line:} Give your analysis results in detail

\noindent \textbf{\textit{4. Important Notes:}}

\noindent 4.1 Please analyze according to the above requirements and respond strictly in the specified format.

\noindent 4.2 Be precise in distance estimation and target identification.

\noindent \sethlcolor{yellow!35}\hl{target description: \{text\_prompt\}}

\noindent \sethlcolor{yellow!35}\hl{observation images: \{img\}}

\noindent Now please start thinking step by step.
\end{framed}

\begin{framed}
\noindent \colorbox{lightblue}{\textcolor{darkblue}{\textbf{\textit{Prompt for Complex Instruction Decompose}}}}
\vspace{0.3cm}

\noindent
You will get a text instruction for long-distance navigation in an indoor environment. Now, your task is to decompose the text instruction into reasonable and clear sub-task goals to help the agent complete the complex navigation task step by step.

\noindent \textbf{\textit{1. Task Description:}}

\noindent All sub-tasks need to be expressed in the form of \sethlcolor{gray!10}\hl{\{move to ...\}}, where the target in \{...\} can be a word or description of an object, or a description of a room area.

\noindent \textbf{\textit{2. Examples:}}

\noindent\textit{Example 1:}
\begin{itemize}
\item \textbf{Text prompt:} "Walk into the hallway and at the top of the stairs turn left and walk into the bedroom. Walk past the right side of the bed and turn right into the closet and all the way through to the bathroom. Stop in front of the toilet in the bathroom."
\item \textbf{Response:}
\begin{enumerate}
\item Move to the \{stairs at the end of the hallway\}
\item Move to the \{bed in the bedroom\}
\item Move to the \{closet\}
\item Move to the \{toilet in the bathroom\}
\end{enumerate}
\end{itemize}

\noindent\textit{Example 2:}
\begin{itemize}
\item \textbf{Text prompt:} ``Go to the wooden stairs. Go up the stairs and go between the couch and the table. Walk into the house through the sliding glass door. Go to the television. Go to the refrigerator. Go to the front of the toaster and stop''
\item \textbf{Response:}
\begin{enumerate}
\item Move to the \{wooden stairs\}
\item Move to the \{area between a couch and a table\}
\item Move to the \{sliding glass door\}
\item Move to the \{television\}
\item Move to the \{refrigerator\}
\item Move to the \{toaster\}
\end{enumerate}
\end{itemize}

\noindent \textbf{\textit{3. Output Requirements:}}

\noindent 3.1 Please planning the following text prompt into sub-goals and respond strictly in the specified format.

\noindent 3.2 Do not include any other information.

\noindent \sethlcolor{yellow!35}\hl{Text prompt: \{text\_prompt\}}
\end{framed}

\begin{framed}
\noindent \colorbox{lightblue}{\textcolor{darkblue}{\textbf{\textit{Prompt for Determining EQA Task’s Waypoints}}}}
\vspace{0.3cm}

\noindent
You will act as an agent to complete the task of assisting scene embodied question answering. I will provide you with a question about the scene. In order to answer this question, we must first navigate to the vicinity of the instance involved in the question.

\noindent \textbf{\textit{1. Task Description:}}

\noindent Now, your task is to analyze and determine the description of the target instance you need to move to based on the current question, which can include some necessary spatial context, such as what type of room this target instance is in and what clear objects exist around it. 

\noindent \textbf{\textit{2. Special Case:}}

\noindent If you think it is difficult to infer the exact target instance for the type of question provided, please output \sethlcolor{gray!10}\hl{``We need to go around and check''}

\noindent \textbf{\textit{3. Output Requirements:}}

\noindent You need to output the description directly without including other analysis and additional response content.
\end{framed}

\setcounter{figure}{0}

\section{Supplementary Figures and Videos}

\paragraph{Supplementary Video Descriptions}

\textbf{Video 1}. Navigation demonstrations of BSC-Nav in simulation environments, showcasing nine successful episodes across object-goal navigation, open-vocabulary navigation, text-instance navigation, and image-instance navigation tasks. Each episode displays egocentric observations, top-down maps, and goal descriptions. In the top-down view, blue squares indicate target locations, with surrounding blue regions representing success zones (within a one-meter radius). These demonstrations illustrate BSC-Nav's efficient general-purpose navigation capabilities.

\noindent\textbf{Video 2}. Long-horizon instruction-following navigation of BSC-Nav in simulation environments, comprising six successful episodes from the VLN-CE R2R benchmark with different instruction complexity. Each episode includes egocentric observations, top-down maps, and human instructions. Blue squares mark target locations, while red squares indicate necessary intermediate waypoints along the instructed path.

\noindent\textbf{Video 3}. Object-goal navigation using BSC-Nav in real-world environments: navigating to a table.

\noindent\textbf{Video 4}. Text-instance navigation using BSC-Nav in real-world environments: navigating to a circular sofa with a small tree in the center.

\noindent\textbf{Video 5}. Text-instance navigation using BSC-Nav in real-world environments: navigating to a lounge area with a soft orange cushion.

\noindent\textbf{Video 6}. Image-instance navigation using BSC-Nav in real-world environments: navigating to an access door (image-specified).

\noindent\textbf{Video 7}.Image-instance navigation using BSC-Nav in real-world environments: navigating to a key model (image-specified).

\noindent\textbf{Video 8}. Mobile manipulation using BSC-Nav in real-world environments, Case 1: clearing a desktop, involving one navigation target.

\noindent\textbf{Video 9}. Mobile manipulation using BSC-Nav in real-world environments, Case 2: relocating a cookie box, involving two navigation targets.

\noindent\textbf{Video 10}.  Mobile manipulation using BSC-Nav in real-world environments, Case 3: preparing oatmeal breakfast, involving three navigation targets.

\begin{figure*}[h]
  \centering
  \includegraphics[width=0.9\linewidth]{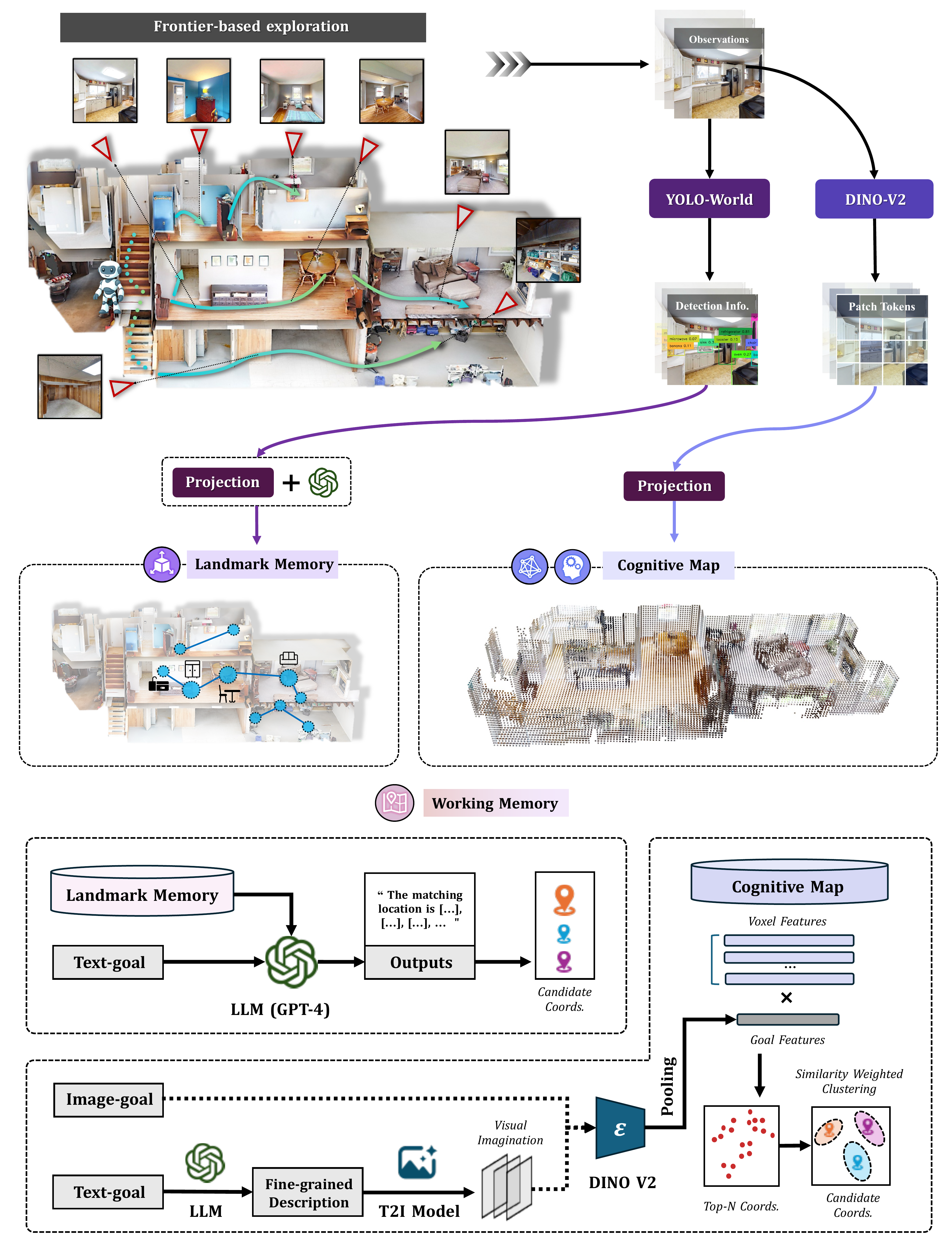}
   \caption{\textbf{Construction and exploitation of structured spatial memory.} 
   BSC-Nav first initializes structured spatial memory through frontier-based exploration strategies. Observational information is processed in parallel, employing YOLO-World for salient object detection and DINO-V2 for patch-level visual embedding extraction, which are respectively used to construct the landmark memory and the cognitive map. During navigation initialization, the working memory module performs hierarchical retrieval based on instruction modality and granularity. For simple textual targets, BSC-Nav employs LLMs for contextual retrieval over the landmark memory.
   For specific textual descriptions and image targets, BSC-Nav performs associative enhanced retrieval over the cognitive map to generate candidate exploration sequences.
   }
    \label{fig:SuppFigure1}
\end{figure*}

\begin{figure*}[t]
  \centering
  \includegraphics[width=0.98\linewidth]{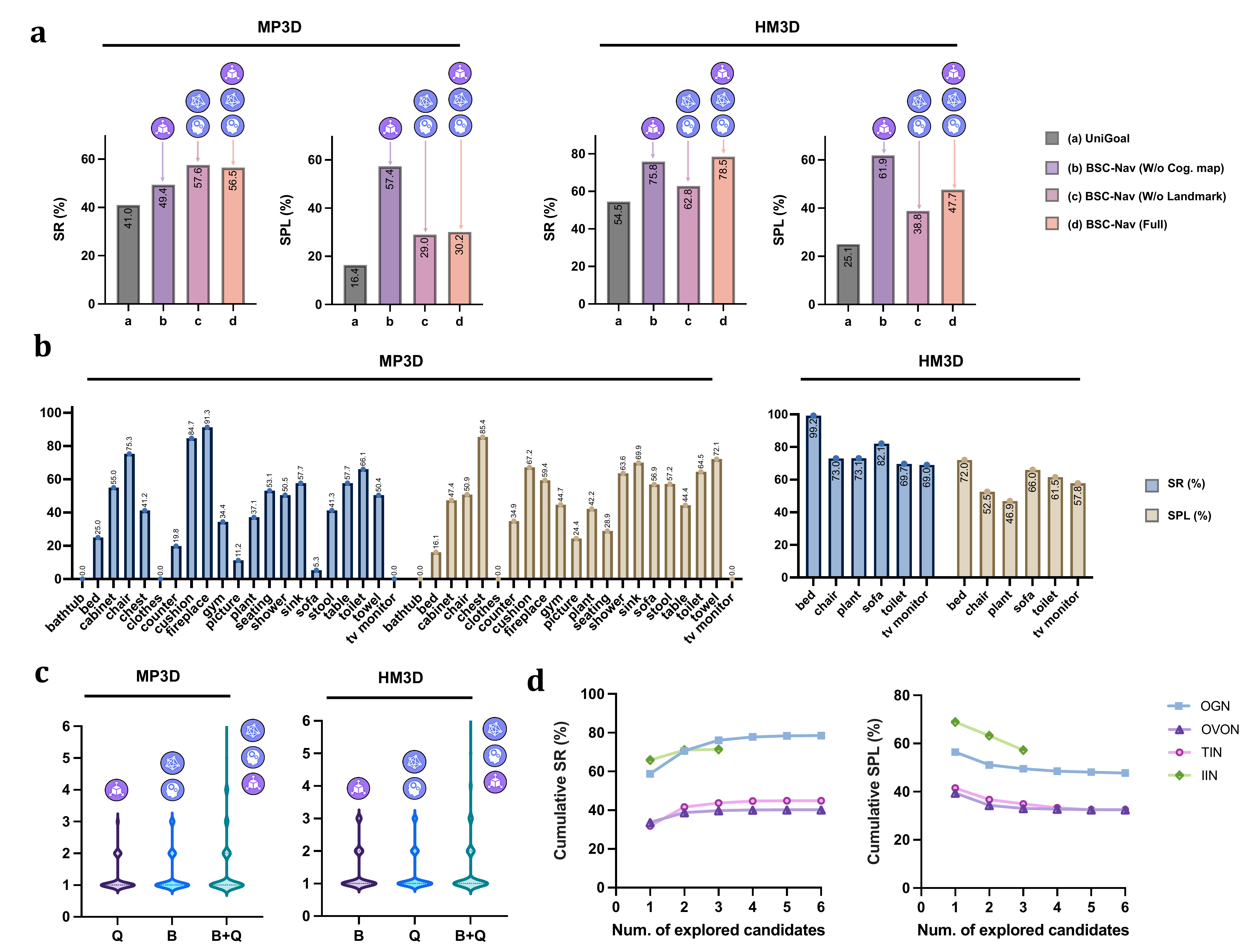}
   \caption{\textbf{Additional results of navigation performance.} 
   \textbf{a}, Comparison of SR and SPL between BSC-Nav and the state-of-the-art Unigoal method in object-goal navigation using landmark memory only, cognitive map only, and both. 
   \textbf{b}, Category-wise SR and SPL across 20 object categories in MP3D and 6 object categories in HM3D. 
   \textbf{c}, Number of candidate coordinates explored during successful navigation episodes on MP3D and HM3D. B, landmark memory retrieval only. Q, cognitive map retrieval only. B+Q, both B and Q. 
   It can be seen that most successful navigation episodes achieve the goal at the first explored coordinate. 
   \textbf{d}, The cumulative SR and cumulative SPL as the number of explored candidate locations increases. Additional explorations improve SR (navigation success) but reduce SPL (navigation efficiency).
   }
   \label{fig:SuppFigure2}
\end{figure*}

% \section{Supplementary Video}

% \subsection{Computational model}
% proof

% too detailed designs / loss function

% \subsection{Experimental details}

% dataset details

% network architectures

% \subsection{Extended results}

\end{document}